\documentclass{article}[verbose=true,letterpaper]
\usepackage{graphicx} 
\usepackage{amsmath}
\usepackage{amssymb}
 \usepackage{booktabs}
 \usepackage{enumitem}   
 \usepackage{array} 

\usepackage[title]{appendix}
\usepackage{hyperref}
\hypersetup{
  pdftitle={Quantifying detection rates},
  pdfsubject={cs.CY, cs.AI},
  pdfauthor={Paolo Bova, Alessandro Di Stefano, The-Anh Han},
  colorlinks=true,       
  linkcolor=blue,       
  citecolor=blue,       
  filecolor=magenta,        
  urlcolor=cyan,        
  linktoc=page,            
  pdfpagemode=FullScreen,
}

\usepackage[
natbib=true,
backend=biber,
style=apa,
sorting=nyt,
minsortnames=1,
maxsortnames=2,
date=year,
labeldate=year,
doi=true
]{biblatex}
\usepackage{arxiv} 
\usepackage[T1]{fontenc}    
\usepackage{subcaption}
\renewcommand{\shorttitle}{Quantifying detection rates for dangerous capabilities}

\usepackage{subcaption}

\title{Quantifying detection rates for dangerous capabilities: a theoretical model of dangerous capability evaluations}
\date{October 2024}

\newtheorem{proposition}{Proposition}
\newtheorem{theorem}{Theorem}[section]
\newtheorem{corollary}{Corollary}[theorem]

\usepackage{authblk}
\usepackage{tikz}
\usepackage{hyperref}

\DeclareRobustCommand{\orcidicon}{
  \begin{tikzpicture}
    \draw[lime, fill=lime] (0,0) circle [radius=0.16] 
      node[white] {\fontfamily{qag}\selectfont \tiny ID};
    \fill[white] (-0.0625,0.095) circle [radius=0.007];
  \end{tikzpicture}
  \hspace{-2mm} 
}

\foreach \x in {A, B, C}{%
  \expandafter\xdef\csname orcid\x\endcsname{\noexpand\href{https://orcid.org/\csname orcidauthor\x\endcsname}{\noexpand\orcidicon}}%
}

\author[1]{Paolo Bova\orcidA{}\thanks{\tt{paolobova@protonmail.com}}}
\author[1]{Alessandro Di Stefano\orcidB{}\thanks{\tt{A.DiStefano@tees.ac.uk}}}
\author[1]{The Anh Han\orcidC{}\thanks{\tt{T.Han@tees.ac.uk}}}

\affil[1]{\href{https://research.tees.ac.uk/}{Teesside University}}

\begin{document}

\maketitle

\begin{abstract}

We present a quantitative model for tracking dangerous AI capabilities over time. Our goal is to help the policy and research community visualise how dangerous capability testing can give us an early warning about approaching AI risks.

We first use the model to provide a novel introduction to dangerous capability testing and how this testing can directly inform policy. Decision makers in AI labs and government often set policy that is sensitive to the estimated danger of AI systems, and may wish to set policies that condition on the crossing of a set threshold for danger. The model helps us to reason about these policy choices.

We then run simulations to illustrate how we might fail to test for dangerous capabilities. To summarise, failures in dangerous capability testing may manifest in two ways: higher bias in our estimates of AI danger, or larger lags in threshold monitoring. We highlight two drivers of these failure modes: uncertainty around dynamics in AI capabilities and competition between frontier AI labs.

Effective AI policy demands that we address these failure modes and their drivers. Even if the optimal targeting of resources is challenging, we show how delays in testing can harm AI policy. We offer preliminary recommendations for building an effective testing ecosystem for dangerous capabilities and advise on a research agenda.

\begin{figure}[ht]
  \centering
  \includegraphics[width=\textwidth]{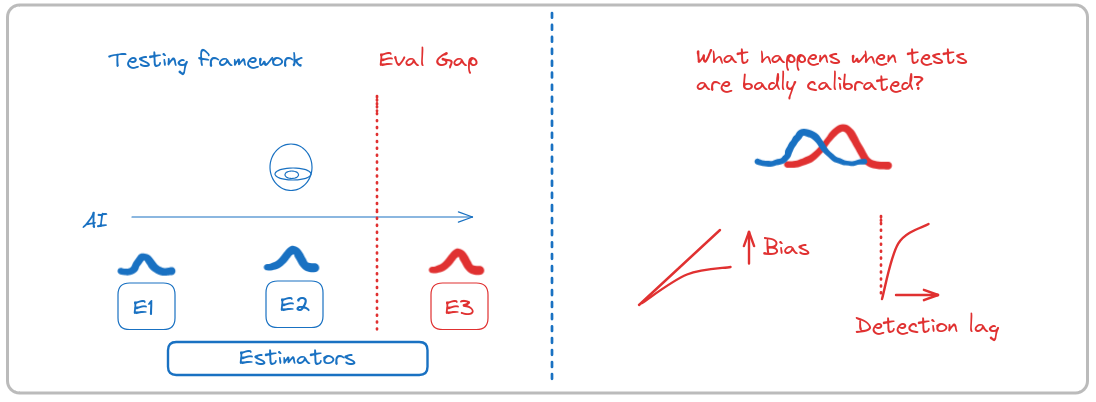}
  \caption{Graphical Abstract: We present a model of testing for dangerous capabilities and show how to model the impact of a gap in evaluations on the quality of our estimator for systemic risks. As AI systems become more intelligent we must improve the calibration of our tests or risk a growing bias in tracking dangerous capabilities and larger lags in detecting the crossing of capability thresholds.}
  \label{fig:graphical_abstract}
\end{figure}

\end{abstract}
\newpage

\section*{Executive summary for general readers}

This executive summary provides an overview of the key concepts and findings suitable for readers who seek a high-level understanding without delving into technical details.

\begin{itemize}
   \item Our goal is to help the policy and research community visualise how successful dangerous capability testing can be in informing us about approaching AI risks.
  \item  We build a model from first principles about one type of information that dangerous capability testing can offer us: an estimate of the lower bound of what dangerous capabilities frontier AI systems are capable of.
  \item We explore the effectiveness of dangerous capability testing across a wide set of basic scenarios for AI evaluations.
  \item We highlight several barriers to dangerous capability testing. These include uncertainty about the dynamics of progress in AI capabilities and competition between frontier AI labs. These barriers work in two ways: either, they eliminate the build-up of testing resources needed for a high-quality estimator, or they prevent the relevant actors from acting on the high-quality estimator.
  \item When building a testing framework incrementally over time, we find that one should balance investment in higher severity tests with tests closer to the current estimated frontier. This allows consistent progress in tracking AI capabilities, while setting a ceiling on lags in threshold monitoring.
\end{itemize}

%
\newpage

\section*{Executive summary for researchers}

This executive summary is designed for readers who are interested in the technical aspects of the study.

\begin{itemize}
  \item We present a model of dangerous capability testing. In this model, the goal of dangerous capability testing is to estimate how dangerous an AI system is.
  \item Assumption 1: Our set of tests can be ordered according to the severity of the dangers they are capable of detecting. We will consider that the severity (or level) of danger can be measured along a single dimension, which we denote by $y$.
  \item Assumption 2:  We can define a "test sensitivity" function $r(y)$ that represents the rate at which we detect an AI system can achieve a level of danger $y$, conditional on ignoring any tests that the system could be more dangerous than $y$.
  \item  Assumption 3: The main estimator of interest applied to the results of an evaluation is the supremum of the danger suggested by the tests. For a given set of tests, denoted by $M$, the event that an AI system achieves a danger level $y$ is represented by $M(y) = 1$. The estimator for the maximum detected danger level can then be expressed as:
\[\hat{y} =  \sup(y: M(y)= 1)\]
\item     Proposition 1: We denote the latent value of $y$ at time $t$ as $y_t$, and further assume that any tests that aim to measure a danger level $y>y_t$ will automatically fail (so we ignore tests in the range $[y_t, y_{max}]$).
    
Given Assumptions 1-3, the cumulative distribution function (CDF) of our estimator $\hat{y}$ can be specified as:
\[
F(\hat{y}) = \exp\left(-\int_{\hat{y}}^{y_t} r(u) \, du\right)
\]
where $r(y)$ is the test sensitivity rate of the distribution. Given the form of the CDF above, we can also call the "test sensitivity" function the reverse-hazard rate or accumulation rate of the estimator distribution.
\item Corrolary 1: Assume $r(y)$ is a piecewise step function. This choice entails a piecewise CDF. If we have $n$ pieces, we could denote the right endpoints of the segments as $0 \leq e_1 < e_2 < \ldots < e_n \leq y_{\text{max}}$.

The CDF for $\hat{y}$ in the $l$-th segment $[e_{l-1}, e_l]$ would then be given by:

\[
F(\hat{y}) = \exp\left(-k_l(e_l - \hat{y}) - \sum_{j > l} k_j(e_j - e_{j-1})\right)
\]

where $k_l$ is the constant reverse-hazard rate in the $l$-th segment.
\item We can compute the bias of the estimator as the model capability increases. Denote the current value of $y$, which tends to increase over time, as $y_t$. Then we can write the bias as:
\[\textbf{Bias} = E[\hat{y}|y_t]  - y_t\]
\item We can compute the likelihood of detecting a crossing of a set dangerous capability threshold. Denote a threshold value of $y$ that we are interested in detecting as $y^*$. Then we can write this detection likelihood as:

\[\textbf{Pr}(\hat{y} \geq y^* | \hat{y}  \leq y_t) = 1 - F(\hat{y} = y^*| \hat{y}  \leq y_t) \]
\item Plotting the estimator bias and threshold detection likelihood for different scenarios is sufficient to capture the results described in the Executive Summary.
\item  This is a tractable model of dangerous capability testing that can be placed in practically any quantative model of AI race dynamics or AI governance (which may be of great interest to economists, forecasters, computer scientists, and other scholars in the field of AI). 

\end{itemize}

\newpage

\section{Introduction}


AI systems are beginning to show increasing levels of dual-use and dangerous capabilities \citep{phuong2024evaluating, park2024ai}. Deception, autonomous R\&D, and assistance with CBNR threat actors are the most well known of these dangerous capabilities as a result of internal and external evaluations of the leading AI models on the frontier \citep{benton2024sabotage, kinniment2024evaluating, institute2024pre1, institute2024pre2}. They are not the only ones that AI safety experts anticipate: a selection of additional risks include multi-agent risks, such as collusion between AI systems, systemic risks such as the shrinking of human-agency, and power-seeking behaviour when combined with long-term planning or strategising \citep{hendrycks2023overview, bengio2024managing}.

With large budgets heading into expanding the AI frontier, the value of information about future dangerous capabilities is large \citep{sevilla2023please}. In comparison, we are seeing very little investment in dangerous capability testing for frontier models. If we severely underestimate dangerous capabilities from AI, then we are likely to be unprepared to respond to any threats posed by misaligned AI systems \citep{hendrycks2022unsolved}.

While regrettable, it is also easy to appreicate why we are neglecting these investments. Dealing with uncertainty is difficult, and it is difficult to know at what rate we are moving towards AI systems with such capabilities.  

We contribute a structured way of thinking about the value of information that dangerous capability testing provides. We build a model from first principles about one type of information that dangerous capability testing can offer us: an estimate of the lower bound of what dangerous capabilities frontier AI systems are capable of. Our goal is to help the policy and research community visualise how dangerous capability testing informs us about approaching AI risks.

We believe that for the first time we have formalized a way to specify dangerous capability tests inside a model of AI development. The model itself is simple and modular, so it can be easily used in other models. As the model is analytically tractable, we can include the model in, say, a model of AI race dynamics without increasing computational complexity. We suspect that this will be of interest to others wishing to model or quantify the impacts of AI policy.

We believe our model can contribute in the following ways: 
\begin{itemize}
    \item help practitioners reason about how their investments aid tracking of AI dangers
    \item  help policymakers visualize the challenges to effective tracking of AI dangers, and potentially help in visualizing progress towards that goal
    \item help think through how AI race dynamics and the policy regime can leave the dangerous capibilities testing ecosystem undeveloped or vulnerable to perverse incentives.
    \item  provides a tractable model of evaluations that can be placed in practically any quantative model of AI race dynamics or AI governance (which may be of great interest to economists, forecasters, computer scientists, and other scholars in the field of AI).
\end{itemize}

In the sections to follow, we first survey the literature on dangerous capability testing, as well as the literature on modeling AI races (Section~\ref{sec:review}). We then give an overview of the model, which discusses the model assumptions and structure, keeping technical discussions to a minimum (Section~\ref{sec:model}). We first illustrate the simplest cases of the model, then move onto a more general discussion of the factors that influence the effectiveness of dangerous capability testing (Section~\ref{sec:results}). We also explore how to build and not to build an AI testing environment over time and describe various scenarios. At the end of Section~\ref{sec:results}, we discuss a potential approach to analysing current AI safety evaluations using our model.

We have only touched the surface of what is possible when modelling dangerous capability testing. In the final section of this paper, we discuss a set of technical research questions that the interested scholar may wish to pursue (Section~\ref{sec:questions}).

\section{Literature Review}
\label{sec:review}

In \citet{shevlane2023model}, one can find an introduction to model evaluations. Tracking AI capabilities is useful not only for AI labs, but for increasing government capacity for governing AI, for a motivation see \citet{whittlestone2021why}. \citet{bengio2024international} and \citet{hendrycks2023overview} provide an overview of topics in AI safety and discuss a range of systemic risks from frontier AI systems.

\citet{gruetzemacher2023international} provide an overview of what the current evaluation ecosystem looks like for frontier AI systems. In short, the ecosystem consists of a handful of specialized organizations, typically NGOs, that test for specific risks from frontier AI systems. Countries have also established their own AI Safety Institutes that conduct similar evaluations on a range of topics, including risks affecting national security. These organisations are also collaborating with each other and private evaluators to coordinate and share knowledge. Recently, an international gathering of AI Safety Institutes was held to help coordinate efforts \citep{t2024fact}. In the current work, we do not explore the international coordination of evaluators further. The interested reader may wish to consult works that propose solutions \citep{ho2023international}.

However, many of the challenges identified by \citet{gruetzemacher2023international} remain to be addressed. Given the voluntary nature of all current AI safety evaluations, race-to-the-bottom dynamics could disincentivise the adoption of strong evaluation methods, and we may fail to scale up the evaluation ecosystem to fulfil the need for high quality evaluations, especially if there is aggressive scaling to create more capable AI systems \citep{sevilla2024can}.

Our work is partly a technical work on how to reconcile various signals about the risks presented by AI systems. Such a discussion could be useful to evidence-based policy making (see \citet{reuel2024open1}, table 1 for an overview of many related open problems in AI governance).

There already exist several qualitative frameworks for using model evaluations in AI policy: the first framework we touch on are safety cases. \citet{buhl2024safety, goemans2024safety} propose that AI developers should make a structured argument with evidence to justify that new AI systems that push the frontier are safe to deploy (similar to safety cases made in the aviation and nuclear industries). Next, we have Responsible Scaling Policies (RSPs) \citep{r2023responsible}. This proposal by METR has developers set thresholds for danger that can be verified to determine if their model exhibits capabilities that the developers consider too risky to train further nor deploy. Evaluations in such a framework become essential evidence that would inform what companies do with their new AI system\footnote{Responsible Scaling Policies (RSPs) have become critical to coporate policy at key labs pursuing Artificial General Intelligence (see the RSPs for Open AI, Antrhopic, and Google Deepmind here) and the Seoul Summit on AI Safety released a declaration where 18 companies agreed to create their own RSPs.}. Note that these are voluntary commitments and there are no explicit consequences for reneging on them\footnote{It is worth noting that while METR has promoted their use as voluntary commitments, they intend for RSPs to serve as a guide to constructing stronger regulation over time \citep{r2023responsible}.}. \citet{koessler2024risk} present an approach to using data on downstream harms to quantify the level of risk that AI systems may pose. They then propose setting a risk threshold: models passing the threshold would pose an intolerable level of risk. These risk thresholds extend the idea of a dangerous capability threshold that we focus on in this paper\footnote{We think the concept of risk thresholds would make for an excellent extension to our work. Yet, the need for additional data on downstream harms to quantify risks means it is currently challenging to adopt. So, we choose to focus on the currently more widely used concept of a dangerous capability threshold.}.  Finally, \citet{dalrymple2024towards} outline a proposal to provide stronger guarantees that future AI systems are safe. Their proposal relies on using technical methods to verify that AI systems do not exhibit harmful behaviours on a sufficiently rich model of the world \footnote{It remains to be seen if there are relatively strong forms of this proposal which could be implemented before the deployment of AI systems. If so, policy could condition deployment or training on strong guarantees of safety.}.

Our work also shares much in common with approaches to modeling the race dynamics between AI companies \citep{armstrong2016racing, han2020regulate, jensen2023industrial}. These works model AI companies as making a tradeoff between safety and the performance of their AI systems and typically find that under a wide range of assumptions that AI companies typically compete by sacrificing safety. 

There have been a number of approaches to modeling how policy can mitigate race dynamics in the AI industry \citep{han2022voluntary, cimpeanu2023social}. Our work focuses on dangerous capability testing, so functions as to clarify the structure of how monitoring AI systems on safety may work in practise. It also may help discriminate between higher and lower quality auditing or monitoring, which can be essential for enabling the evolution of safe development behaviours  \citep{bova2024both}.

\section{Model}
\label{sec:model}
We present a model of dangerous capability testing. In this model, the goal of dangerous capability testing is to estimate how dangerous an AI system is. \footnote{Organisations who do dangerous capability testing often have other goals, such as learning why AI systems are more dangerous, predicting future potential sources of risk, or investigating methods for mitigating dangers. We do not directly model these considerations in the present work, though they are important enough for further discussion, see later sections.}

For the sake of legibility, we will refer to dangerous capability testing as evals. This shorthand is common in the AI industry where it can be used to refer to a wide range of capability and safety tests; the phrase model evals is frequently used in the same contexts.

We will also frequently shorten "dangerous capabilities" to dangers. Note that we do not explicity model the harms that AI systems with dangerous capabilities can cause (see \citet{hendrycks2023overview} for an overview). When deriving the model, we also do not specify the nature of the dangerous capabilities being tested. We argue that one strength of our model lies in seperating the process of dangerous capability testing from discussions of the capabilities and harms themselves.\footnote{With that assumption made clear, readers may be intereted in a later section where we discuss how our model would be affected by dangerous capabilities that make AI systems resistant to effective testing.}

Here, we define evals as a set of tests, $M$, built and run by an auditor or lab to estimate the level of danger of one or more AI systems. It is useful to recognise that tests will usually be imperfect, imprecise, and incomplete (although we will also be able to model scenarios where tests are practically perfect). 

This inherent uncertainty in evals leads us to propose the following two propositions:

\begin{proposition}
    Our set of tests can be ordered according to the severity of the dangers they are capable of detecting.\footnote{ In the case that tests are overlapping, we can think of the overlapping testing region as a composite test that is capable of detecting its own level of danger.} We will consider that the severity of danger can be measured along a single dimension, which we denote by $y$.
\end{proposition}

\begin{proposition}
   We can define a "test sensitivity" function $r(y)$ that represents the rate at which we detect an AI system can achieve a level of danger $y$, conditional on ignoring any tests that the system could be more dangerous than $y$.
\end{proposition}

As we derive our model for evals based on first principles, these two propositions are primarily responsible for the final model we end up with. Briefly, we note that organisations conducting evals and AI Safety researchers do indicate that more severe dangerous capabilities are likely harder for AI systems to achieve and expect to test for them differently. This provides tentative support for Proposition 1. Proposition 2 suggests that evaluators can judge to some degree that some levels of danger are better tested for than other levels. This is typically a difficult undertaking in a nascent and experimental field, but as this is crucial for informing any model of how to allocate investments in safety evals, we later present some arguments for how to estimate the relative sensitivity of different tests. 

Combining the two propositions above, we can describe our set of tests as a function that maps the level of danger we test for to the rate of detection. In statistics, this function would be akin to a reverse-hazard rate or accumulation rate. To better captuing the meaning of this function in our use case, we will refer this function as the test sensitivity rate. Intuitvely, if we want to have a high chance of correctly estimating the level of danger of the AI system, we need our test sensitivity rate to be high for all values of danger it makes sense to test for. 

\subsection{Estimator Distribution}

Above, we proposed to model the inherent uncertainty in evals as a test sensitivity function. We now show that due to this uncertainty, we can describe the result of a set of evals by a probability distribution.

In statistics, it is common to propose an estimator, $\hat{y}$, for a result of interest. Evals are usually interested in an estimator for the highest level of danger we find an AI system is capable of. In other words, we often seek a lower bound on how dangerous an AI system is by looking at the highest level of danger implied by the evaluation.

\begin{proposition}
    The main estimator of interest applied to the results of an eval is the supremum of the danger suggested by the tests. Given a set of tests, $M$, and denoting the event of detecting an AI system achieving danger severity $y$ by $M(y) = 1$, this estimator can be written as:
\[\hat{y} =  \sup(y: M(y)= 1).\]
\end{proposition}
Note that our choice of estimator is another crucial choice in determining our model of evals. If we were mainly interested in a different estimator, for example the median level of danger presented by an AI system, then the distribution would have a different form. Still, the current focus of organisations conducting AI safety evals is to identify the emergence of dangerous capabilities as early as possible \citep{r2024details, benton2024sabotage}. This motivates the choice of estimator we have chosen here.\footnote{At this stage, we note that while it is possible to use similar methods to derive a lower bound estimator for what AI systems cannot do (i.e. an infinum), which is also of interest to these organisations, we leave the analyis of such an estimator to future work.}

The estimator $\hat{y}$ is a random variable and, therefore, follows a probability distribution, which we will denote as $F$. 
Formally, the cumulative density function (CDF) denoted $F$ is the likelihood that our estimator $\hat{y}$ is less than or equal to a given value $y$, which we can write succinctly as $F(\hat{y}) = P(\hat{y} \leq y)$.

Whenever we know $F$, we can answer the following statistical questions about current and future evals:

\begin{itemize}
    \item How biased can we expect our estimator for danger to be?
    \item How efficient is our estimator?\footnote{In the form of distribution I propose, we will not immediately be able to answer this question, so we will revisit efficiency in future work.}
    \item  For a given threshold for danger
    \begin{itemize}
        \item How likely are we to miss AI systems that cross this threhsold?
        \item How large should we expect the lag time (in units of AI progress) to be before we detect the crossing?
        \item How large will the bias be when we reach the threshold?
    \end{itemize}
\end{itemize}

For reasons we will describe in more detail in Section~\ref{sec:case_study}, quantifying $F$ accurately will usually be challenging. Nevertheless, we believe there is much to learn from understanding our estimator, even if the available evidence only helps us capture the relative efficacy of evals for different levels of danger.

We can derive an explicit form for $F$ from first principles (see Section~\ref{sec:proof1} for a proof).

\begin{theorem}
    We denote the latent value of $y$ at time $t$ as $y_t$, and further assume that any tests that aim to measure a danger level $y>y_t$ will automatically fail (so we ignore tests in the range $[y_t, y_{max}]$).
    
Given Propositions 1-3, the cumulative distribution function (CDF) of our estimator $\hat{y}$ can be specified as:
\[
F(\hat{y}) = \exp\left(-\int_{\hat{y}}^{y_t} r(u) \, du\right),
\]
where $r(y)$ is the test sensitivity rate of the distribution. Given the form of the CDF above, we can also call the "test sensitivity" function the reverse-hazard rate or accumulation rate of the estimator distribution.
\label{theorem1}
\end{theorem}

This concept of a reverse-hazard rate is likely familiar only to those with a background in survival analysis,  a branch of statistics that usually analyzes the expected duration of time until an event occurs (such as mechanical failure, a disaster, or biological death). This branch of statistics is relevant to our problem: we can frame our work as the analysis of the expected number of AI improvements until the detection of a model's current capabilities.

Although the patterns of results we find typically hold for more than one choice of $r(y)$, for the remainder of this paper, we will specify that $r(y)$ is a (usually) piecewise step function of the danger. 

\begin{corollary}
Assume $r(y)$ is a piecewise step function. This choice entails a piecewise CDF. If we have $n$ pieces, we could denote the right endpoints of the segments as$0 \leq e_1 < e_2 < \ldots < e_n \leq y_{\text{max}}$.

The CDF for $\hat{y}$ in the $l$-th segment $[e_{l-1}, e_l]$ would then be given by:

\[
F(\hat{y}) = \exp\left(-k_l(e_l - \hat{y}) - \sum_{j > l} k_j(e_j - e_{j-1})\right),
\]

where $k_l$ is the constant reverse-hazard rate in the $l$-th segment.
\label{corollary1}
\end{corollary}

\subsection{Measures of estimator effectiveness}

Now that we have defined our estimator, we can define and derive the following outcome variables to help us think critically and broadly about the value of dangerous capability testing:

\begin{enumerate}
    \item The bias of the estimator as model capability rises (how badly we fail to track model dangers as capabilities grow).

    Denote the current value of $y$, which tends to increase over time, as $y_t$. Then we can write the bias as:
\[\textbf{Bias} = E[\hat{y}|y_t]  - y_t\]

    Note that any tests for $\hat{y}> y_t$ are assumed to automatically fail (and so have reverse-hazard rate of $0$). So the $\textbf{Bias}$ will be positive.\footnote{We made this assumption for reasons of tractability, so in the field we may want to consider the possibility of a negative bias due to false positives. Weakening this assumption would be a useful extension to our work.}


    \item The likelihood of detecting a crossing of a set dangerous capability threshold (How likely are we to detect an unacceptably dangerous model as model capabilities grow).

    Denote a threshold value of $y$ that we are interested in detecting as $y^*$. Then we can write this detection likelihood as:
    \[\textbf{Pr}(\hat{y} \geq y^* | \hat{y}  \leq y_t) = 1 - F(\hat{y} = y^*| \hat{y}  \leq y_t). \]
    
    This is trivially $0$ if $y_t < y^*$ and we will show that it is increasing in $y_t$ after $y_t$ exceeds $y^*$. Knowing this detection rate is useful for capturing how good a job our estimator does at alerting us to different excesses beyond the threshold.
    
    The above outcomes are intended to be thought of as a trend that depends on increasing model capabilities. However, it is also informative to reduce these trends down to summary statistics of the effectiveness of dangerous capability testing:

    \item The expected lag time of the estimator in detecting a crossing of a danger threshold (as a summary of how late we might be in detecting a model we believe is too dangerous to deploy).

    To calculate this we need to know how $y_t$ increases over time. Let $t_{lag}$ denote the delay time after which the crossing of threshold $y^*$ is first detected, i.e. the first time we see the estimator exceed the threshold $\hat{y} > y^*$.  Then, for the distribution presented in \ref{corollary1}, the expected lag time, conditional on detecting the AI system crossing the threshold at all, is:
    \[E_S[t_{lag}] = \frac{1}{1-C} \sum_{ l^* \leq j \leq l_{max}} \int_{t_{e_{l-1}}}^{t_{e_l}}t_{lag} s(t_{lag}) dt_{lag},\]
    where $C$ is the probability of completely missing the threshold crossing: \[C = \exp\left(-k_{l^*}(e_{l^*} - y^*) - \sum_{ l^* < j \leq l_{max}} {k_j(e_j - e_{j-1})} \right).\]
    Typically, we find that $C$ varies much more when varying the test sensitivity function than the expected lag time, but we still find it informative to consider both.

\end{enumerate}

\subsection{Dynamics in testing AI systems}
\label{sec:dynamics}

We still need the following underpinnings to present the dynamics of AI evals:  incremental testing of evals and
a production function for evals of different severity, to be elaborated below.

\subsubsection{Incremental testing}

If in each time step, we sample an estimate from our estimator, what information do we carry over to the next time step? It would be a mistake to think we are starting from a blank slate.

For simplicity, let's imagine we never change the suite of tests. However, following Theorem \ref{theorem1}, we know that, as $y_t$ advances, fewer of our high danger tests will automatically fail. Let's further assume that, as long as the tests do not change, that test results of the same kind are persistent after each advance in AI capabilities. This means that by $y_{t+1}$, we already know the results of all tests before $y_t$. Only the tests in between $y_t$ and $y_{t+1}$ provide new information. We can think of how our estimate $\hat{y}$ updates in this setting as follows:

First, if none of the newly applicable tests detect any danger, then the highest severity test which will pass will be the test at $\hat{y}_t$. The probability of this happening is $1 - (F(y_{t+1}) - F(y_t)) = F(y_t)$ since the estimator distribution at $t+1$ is truncated at $y_{t+1}$.  

Second, with likelihood $1 - F(y_t)$, we sample from $[y_t, y_{t+1}]$ with the normalized density $\frac{f(y)}{1 - F(y_t)}$. \footnote{We can take an alternative approach to updating our danger estimate. Assuming we have ruled out false positives, we can say that our current estimate $\hat{y}_t$ is a lower bound on how dangerous the model was last period. However, we are not confident that the new AI system will be pass all applicable tetsts, including those in the range $[\hat{y}_t, y_t]$. This time we left-truncate the distribution at $\hat{y}_t$. With likelihood $F(\hat{y}_t)$, we stick to $\hat{y}_t$. Otherwise we sample from $[\hat{y}_t, y_{t+1}]$ with the noramlized density $\frac{f(y)}{1 - F(\hat{y_t})}$. While this is a fairly natural approach to updating our danger estimate, we usually go for the previous approach since it simplifies the calculation and interpretation of metrics like the expected time lag and the likelihood of missing the threshold crossing.}

What then do we do once we add on new tests? So far, our assumptions mean that our danger estimate is non-decreasing. This means we can ignore any new tests at $t+1$ which target a severity below $\hat{y}_t$. Now assume that whenever we add new tests at a level of severity $y$, that the results of these tests are independent of the results of the old tests at this level. Let the lowest level of $y \in [\hat{y}_t, y_{t+1}]$ affected by the new tests be $\tilde{y}$. Then, the update dynamic is the same as before. With probability $F_{t+1}(\tilde{y})$ we keep estimate $\hat{y}_t$, otherwise we sample from $[\tilde{y}, y_{t+1}]$ with the new normalized density $\frac{f_{t+1}(y)}{1 - F_{t+1}(\tilde{y})}$.\footnote{In the alternative approach, we can replace $\tilde{y}$ with $\hat{y}_t$ since all the tests above $\hat{y}_t$ are still appplicable, not just those that have been changed.}



\subsubsection{A production function for evals of different severity}

Let's consider that investment in test sensitivity is a standard affair that requires only a certain proportion of time spent by a reasonably large research team. A straightforward approach would be to model this as a linear test production function that allows us to reach a relatively consistent rate of detection with a similar number of resources per test interval. In this case, we can predictably move resources from lower end tests to new ones. This is a relatively simple scenario where the question of balancing resources is a bit easier to visualise.


We don't explicitly use the dynamics of incremental testing or the production function in several of the results to follow. However, it is useful to be aware of these assumptions to understand how we ran the simulations on which some of our later figures are based.




\section{Results}
\label{sec:results}

\subsection{An illustration using 1 or 2 test blocks}

\subsubsection{A single test block}

It is helpful to introduce what the model tells us by introducing one test block at a time. Please note that the numbers used here are purely for the purpose of illustration.

\begin{figure}[ht]
  \centering
  \begin{subfigure}{\textwidth}
  \centering
    \includegraphics[width=0.4\linewidth]{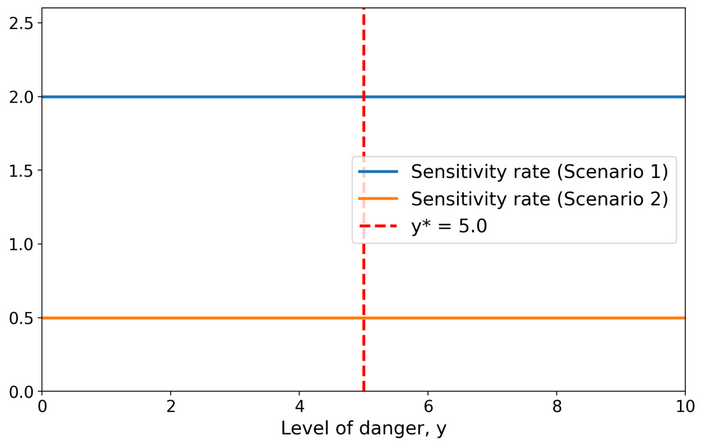}
    \caption{Test sensitivity rate}
    \label{fig:panel1_sub1}
  \end{subfigure}
  \begin{subfigure}{0.4\textwidth}
    \includegraphics[width=\linewidth]{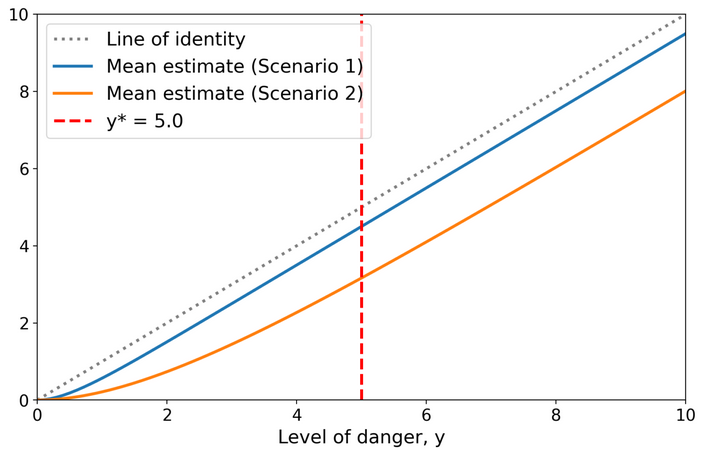}
    \caption{Danger bias}
    \label{fig:panel1_sub2}
  \end{subfigure}
  \begin{subfigure}{0.4\textwidth}
    \includegraphics[width=\linewidth]{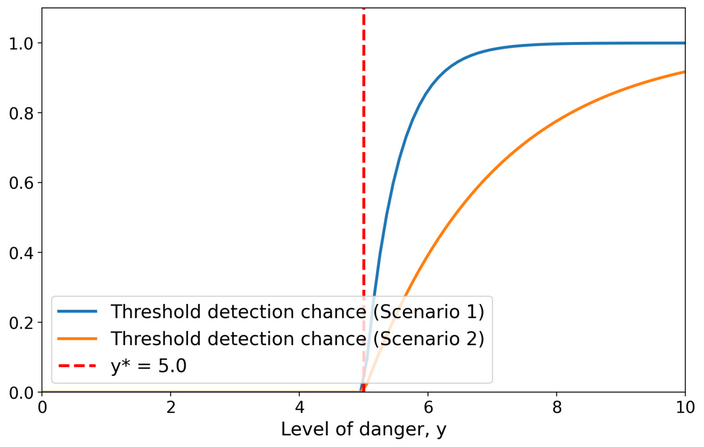}
    \caption{Threshold detection likelihood}
    \label{fig:panel1_sub3}
  \end{subfigure}
  \caption{Single test block: (a) Test sensitivity function for the case of 1 test block: The blue and orange scenarios illustrate different detection rates. The red dashed line indicates our threshold for danger. (b) Bias in our estimator as we vary hidden AI capabilities: Two scenarios are shown. The blue scenario reflects a consistently higher detection rate than the orange scenario as it is much closer to the line $y=x$. The bias in each scenario is the distance between the relevant solid line and the line $y=x$, i.e. $Bias = E[\hat{y}] - y_t$.  (c) Likelihood of detecting a crossing of the danger threshold as AI system capabilities increase. The red dashed line, $y^*$ indicates the danger threshold. By construction, there is $0$ chance of detecting a crossing that has not happened. }
  \label{fig:panel1}
\end{figure}

When we say that an eval consists of one test block, we mean that for the range of danger tested for in the eval, $y_t \in [0, 10]$, the detection rate is constant throughout. We plot this case for two different detection rates in Figure~\ref{fig:panel1}. The blue scenario has a consistently higher detection rate than the orange scenario. Also note that in this scenario the danger threshold $y^*$ happens to be in the middle of the testing block, at $y_t = 5$. We have set this threshold to half of our specified maximum level of danger $y_t=10$.

We can now plot our measures of estimator effectiveness as we vary the hidden dangerous capabilities of new AI systems, $y_t$, see Figure~\ref{fig:panel1}.

    Let's start with the bias in the estimator. In Figure~\ref{fig:panel1_sub2} we plot the bias in our estimator of the lower bound of AI dangers as AI systems become more dangerous. The bias is the distance between the relevant curve and the line of identity $y=x$. As expected, in the blue scenario where the detection rate is consistently higher, the mean estimate of the danger is significantly closer to the true level of danger presented by AI systems. Trivially, the bias of the estimator falls with more effective testing.

The curves in both scenarios quickly become linear as AI dangers increase representing an approximately constant bias (see, for example, the bias in both scenarios at the danger threshold). This matches well with our constant test sensitivity function. We will later see that this bias may change abruptly if our test sensitivity function is not constant.\footnote{The bias actually increases over time despite our use of a constant test sensitivity function. Yet, as AI systems become more capable, the bias quickly approaches a constant depending on our single detection rate, $\frac{1}{k}$. Even in this simple setting, keeping track of the lower bound of AI dangers is an activity that may start off easy, but quickly becomes more difficult over time. However, we advise against updating too strongly on this pattern as often by the time we reach the danger thresholds we set, the bias has effectively converged to the constant $\frac{1}{k}$. This is driven by the nature of our estimator: as we truncate $\hat{y}$ at $0$, the mean initially places more mass at $0$, which is of course closer to the true value $y_t$ than any negative number. However, as $y_t$ increases and more tests become relevant, we place less probability mass on $0$ and more on strictly positive numbers. Hence, the bias increases up to a point where the probability mass on $0$ effectively vanishes.}

The second trend of interest to us relates to the likelihood of detecting an AI system crossing the danger threshold we set; see Figure~\ref{fig:panel1_sub3}.

Clearly, a consistently higher detection rate is an advantage here. It may surprise the reader to see how much of an advantage it is. The blue scenario is much more likely to detect the threshold crossing earlier than the orange scenario.

It is possible to summarise this observation using the expected lag time. Assuming AI dangers increase by $1$ per time step, then the expected lag time in scenario 1 is just $0.5$, three times lower than the expected lag time in scenario 2 of $1.5$ (these expected lag times are conditional on detecting the crossing at all, although that is hardly an issue in the scenarios presented in the current example). While this is still less than the four times increase in the detection rate for scenario 1 relative to scenario 2, this increase is still substantial.

It may also be surprising to see that despite having a low detection rate, scenario 2 still has such a good chance of finding out about the crossing. The likelihood of not detecting the crossing at all is just under $0.1$ in scenario 2 and we might have expected the lag time to be much higher.

Our scenarios owe this strong performance to the consistency with which we assume that they test their models.

\subsubsection{Introducing a second test block}

Now that we have introduced the basics of analysing our model, it is time to investigate what happens as we limit testing. Consider that instead of a consistent detection rate, the detection rate falls to $0.1$ suddenly at $y_t=6$ (perhaps evaluators did not have the resources to design and run new tests for these dangers).

\begin{figure}[ht]
  \centering
  \begin{subfigure}{\textwidth}
  \centering
    \includegraphics[width=0.4\linewidth]{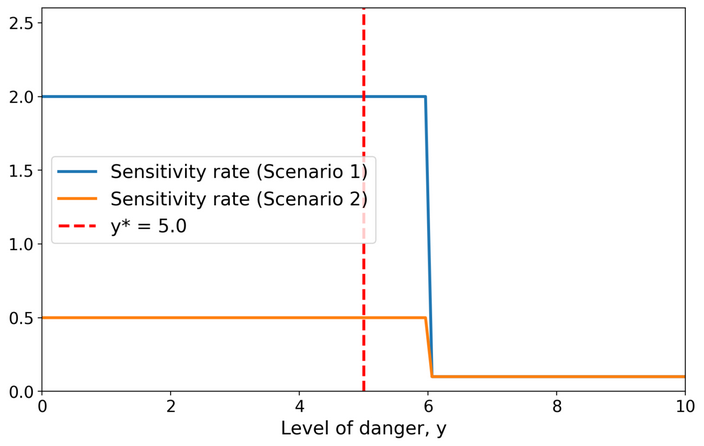}
    \caption{Test sensitivity rate}
    \label{fig:panel2_sub1}
  \end{subfigure}
  \begin{subfigure}{0.4\textwidth}
    \includegraphics[width=\linewidth]{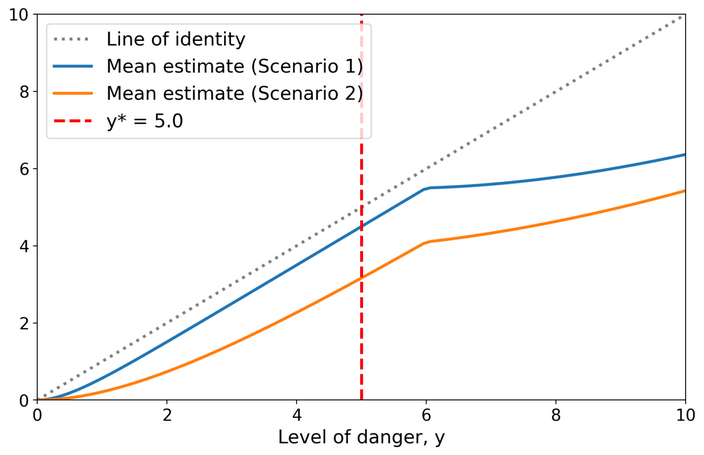}
    \caption{Danger bias}
    \label{fig:panel2_sub2}
  \end{subfigure}
  \begin{subfigure}{0.4\textwidth}
    \includegraphics[width=\linewidth]{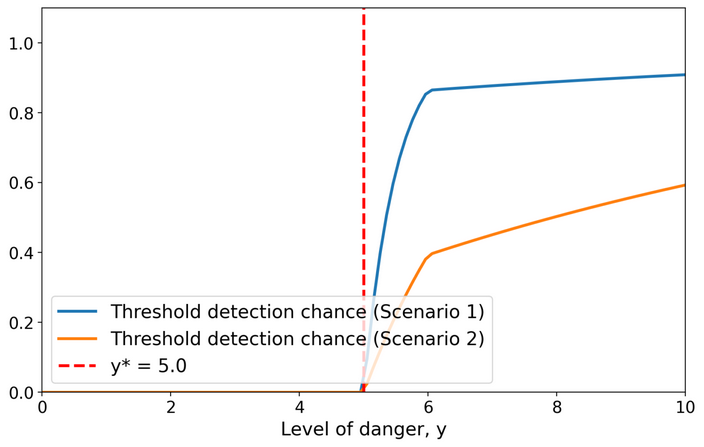}
    \caption{Threshold detection likelihood}
    \label{fig:panel2_sub3}
  \end{subfigure}
  \caption{Two test block: (a) Test sensitivity function for the case of 1 test block that misses many dangers: The blue and orange scenarios illustrate different detection rates. The red dashed line indicates our threshold for danger.(b) Bias in lower bound estimator as we vary hidden AI capabilities when there are limits for what dangers we can test for: The bias in each scenario is the distance between the relevant solid line and the line $y=x$, i.e. $Bias = E[\hat{y}] - y_t$.  (c) Likelihood of detecting a crossing of the danger threshold as AI system capabilities increase when we have limits in what dangers we can test for past $y_t=6$. The red dashed line, $y^*$ indicates the danger threshold. By construction, there is $0$ chance of detecting a crossing that has not happened.}
  \label{fig:panel2}
\end{figure}

As expected, the bias starts to increase dramatically past $y=6$ in both scenarios.

What is more interesting is that the differences between the two scenarios become more pronounced for the detection of a crossing of the danger threshold. While scenario 1 still does quite well with an expected lag time of just under $0.5$ and a non-detection probability of around $0.1$, scenario 2 now does much worse, with a $0.4$ chance of missing the crossing altogether. Even conditional on detecting the crossing, scenario 2 still expects a lag of 1.27.\footnote{While lower than before, remember this is conditional on detecting the crossing and there is now a 40\% compared to a 10\% chance of missing the crossing.}

\begin{figure}[ht]
  \centering
  \begin{subfigure}{\textwidth}
  \centering
    \includegraphics[width=0.4\linewidth]{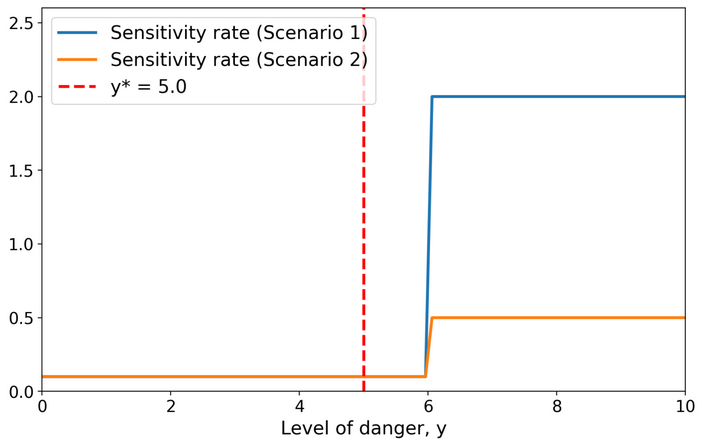}
    \caption{Test sensitivity rate}
    \label{fig:additional_cases_sub1}
  \end{subfigure}
  \begin{subfigure}{0.4\textwidth}
    \includegraphics[width=\linewidth]{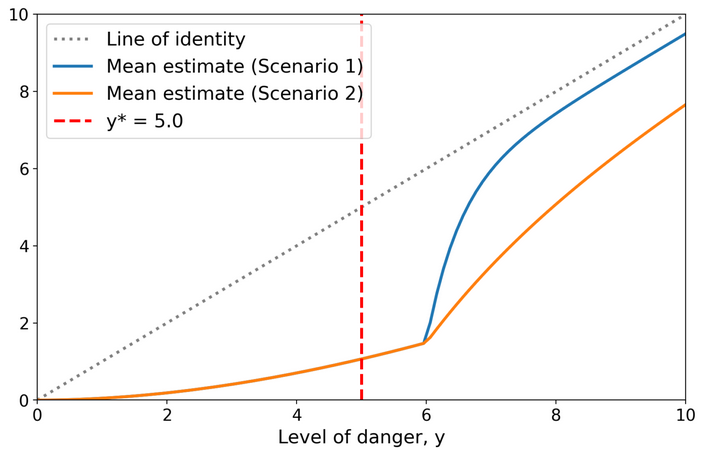}
    \caption{Danger bias}
    \label{fig:additional_cases_sub2}
  \end{subfigure}
  \begin{subfigure}{0.4\textwidth}
    \includegraphics[width=\linewidth]{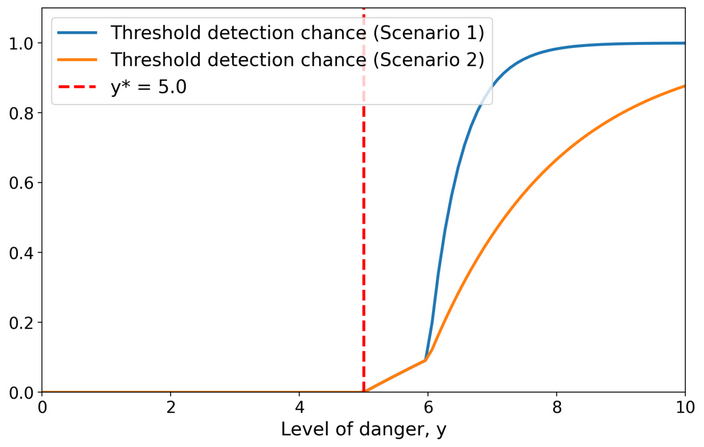}
    \caption{Threshold detection likelihood}
    \label{fig:additional_cases_sub3}
  \end{subfigure}
  \caption{Impact of reversing test sensitivity rate on bias and detection likelihood.}
  \label{fig:additional_cases}
\end{figure}

\subsubsection{What happens if we reverse the detection rates?}

We might be interested in what happens if we focus all our testing efforts beyond the danger threshold, essentially swapping the detection rates of the first and second segments in each scenario; see Figure~\ref{fig:additional_cases_sub1}.

As we are no longer testing much for low end dangers, we can see a large growing bias in our lower bound for AI danger as models get more dangerous. By the time we reach our danger threshold, the bias is incredibly large, Figure~\ref{fig:additional_cases_sub2}.  Once we start testing again past $y=6$ we can reduce this bias very quickly if the detection rate is high enough (in scenario 1 but not scenario 2). Even still, it is not clear if this is a favourable scenario to be in as it is quite likely that dangerous capabilities have taken us by surprise.

A similar story is visible in Figure~\ref{fig:additional_cases_sub3}. Our chance of detecting the crossing is very low until we start testing again after $y=6$.  A high detection rate later on can help us recover a good chance of eventually detecting the crossing, but the expected lag time does increase dramatically. Unsurprisingly, the expected lag time increases by $\approx1$ relative to Figure~\ref{fig:panel1_sub3} due to the very low detection rates between danger levels $5$ and $6$.

The poor performance in detecting the threshold crossing can be repaired if we also tested to a high degree immediately after the threshold (restoring Figure~\ref{fig:panel1_sub3}). However, we would still do a very poor job of tracking the danger level of AI systems before the threshold, potentially leaving us unprepared to tackle the risks once we do detect such systems passing the danger threshold.

\subsubsection{What happens if there is a gap in the middle between two test blocks?}

Let us consider that we have two high-sensitivity testing blocks, but they are separated by some interval where no testing is done at all. The bias over time will resemble Figure~\ref{fig:additional_cases_sub2}, except that the early bias remains relatively small and stable (see the beginning of the x-axis for Figure~\ref{fig:panel1_sub2}). Assuming that the the missing test block contains the danger threshold, we can still use Figure~\ref{fig:additional_cases_sub3} to understand how threshold detection changes over time. All that happens is that the likelihood only starts increasing from $0$ after reaching the new test block, rather than increasing right after the danger threshold. Even though we can eventually recover a high detection rate sometime after the danger threshold is crossed, we may be more likely to be caught by surprise.

\subsubsection{What happens if we change the danger threshold?}

\begin{figure}
  \centering
    \includegraphics[width=0.4\linewidth]{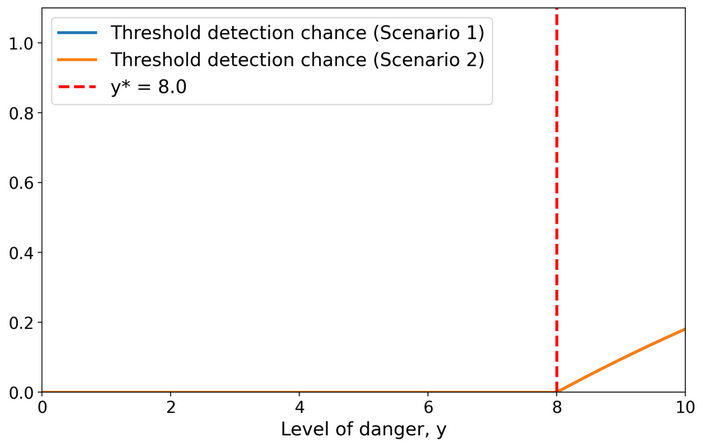}
    \caption{Threshold detection likelihood when setting a higher threshold}
    \label{fig:fig5}
  \end{figure}

Naturally, where we set the danger threshold matters. If we set the danger threshold higher, say past the point where our testing drops off, then we do terribly in both scenarios (while the conditional expected lag time is still around $1$, the chance of missing the crossing increases past $80\%$, see Figure~\ref{fig:fig5}).

If we instead set the danger threshold lower, we have more opportunities to detect the crossing and so are much less likely to miss the threshold. We see an image similar to Figure~\ref{fig:panel1_sub3}, except that the likelihood of detecting the crossing instead begins to increase at the earlier threshold (the shape of each curve remains the same).

\subsubsection{Summary}

We have now explored a wide set of basic scenarios for AI evals. We can summarise the two failure modes we have seen as follows:
  \begin{itemize}
\item Failure Mode 1 - The bias in estimating the danger posed by AI can change much faster at higher levels of AI capability.
      \item Failure Mode 2 - Large lags in threshold monitoring occur when testing is focused before but not after the threshold.
  \end{itemize}

The illustrations we have seen suggest that there can be a balance to be struck when allocating resources between tests that target lower and higher levels of AI danger. For example, attempts to address failure mode 2 (as in Figure~\ref{fig:additional_cases}) may be insufficient to address failure mode 1.

In the next sections, we turn our attention to the barriers facing test effectiveness that lead to these failure modes.

\subsection{Barriers facing test effectiveness}

As illustrated above, a fall in the effectiveness of dangerous capability testing may manifest itself in two failure modes: a higher bias in our estimates of AI danger over time or larger lags in threshold monitoring.

We now present a range of barriers to effective testing. For each barrier, Table~\ref{tab:barriers} discusses how they cause or amplify our two failure modes. These barriers work in two ways: either they eliminate the build-up of testing resources needed for a high-quality estimator, or they prevent the relevant actors from acting on the high-quality estimator. We discuss two of these barriers in more detail. First, we focus on market dynamics that could threaten to eliminate effective safety testing. Second, we discuss reasons for dangerous capability tests becoming more difficult to design and run over time.

Understanding these barriers is crucial to understanding why high-quality testing of novel AI systems is difficult to guarantee. For the sake of brevity, we have omitted discussions of important barriers that have received better treatment elsewhere and are less directly relevant to the proposed model. These include challenges in coordinating AI evaluators \citep{gruetzemacher2023international}, calibration of specific estimators \citep{h_ojmark2024analyzing1}, and strategic underperformance on tests \citep{benton2024sabotage}.

\renewcommand{\arraystretch}{1.5}
\begin{table}[h]
    \centering
    \begin{tabular}{p{4cm}p{4cm}p{4cm}}
        \toprule
        Barrier& Description & Effect on test quality \\
        \midrule
        Market dynamics: competition& Competitive market pressure allocates most resources to getting AI capabilities deployed faster to obtain first mover advantage.& Bottlenecks on test sensitivity at higher levels of danger. Bias increases over time.\\
        Market dynamics: Trends in AI progress& AI labs facilitate orders of magnitude of funding or compute more quickly and this continues a trend of AI capabilities improving.& Less time to build tests for dangerous models, less need for intermediate tests. Sudden shifts in bias more common.\\
        Technical challenges to testing for greater danger& Some dangerous capabilities may be difficult to test for if AI systems have incentives to underperform.& Test sensitivty falls at higher levels of danger. Bias may remain low until high levels of danger reached.\\
        Changes in the relationship between AI progress and danger& Level of dangerous capabilities could follow a "s" shape as generic AI capabilities progress.& Unprepared for testing the most dangerous systems. Bias increases quickly and suprisingly, perhaps without any indication.\\
        Motivated reasoning& Initial warnings of danger are dismissed as false positives, and marginal effort is dedicated to justifying the dismissal (common against a tight deadline).&Despite a short detection lag, AI system is treated as safer than it is. Estimator for danger may be polluted by new tests motivated to justify the dismissal.\\
        Overconfidence in safety& Risks are reported only after applying safety mitigations. &Despite the infancy of these mitigations, AI systems are presented as achieving low levels of risk and further training goes ahead no matter the pre-mitigation risk level.\\
    \end{tabular}
    \caption{Barriers to effective testing.}
    \label{tab:barriers}
\end{table}

\subsubsection{Market dynamics}

Market dynamics can erode initial attempts to invest in a testing ecosystem. 

When we discuss market dynamics in this paper, we refer to two trends: first, total investments in training new AI systems have risen rapidly to exploit new capabilities and predictable scaling laws. Second, we observe that companies compete fiercely with each other to bring more powerful models to market first.

There is a modelling literature that covers tech races to bring a new technology to market first. In such models, there is a large first mover advantage in deploying your systems first \citep{askell2019role, armstrong2016racing, han2020regulate}. So, these models predict that developers will reduce their investment in the safety of their AI systems over time until they reach some low level. The large incentives to be first leads developers to take huge risks. Competition with other developers means that developers have a lot to lose by delaying their plans, leaving limited time for safety tests or mitigations. There is tentative evidence that these dynamics are at play in the current market for generative AI systems. We can see from the data in \citet{cottier2023who} that companies compete with each other to develop more powerful AI systems.\footnote{ \citet{sevilla2024can} build on this work to suggest that by 2030 AI systems could be around $10^4$ times larger than GPT4. Were these trends to hold, and if larger models unlock new capabilities with potentially dangerous consequences, then the outlook for AI Safety, both in design and evaluation, looks poor.} Meanwhile, leading AI labs typically give limited time for external evaluations of their new systems \citep{i2024openai}.

With the above background findings in mind, we now explain how market dynamics can lead to a reduction in new investments for testing the safety of powerful AI systems.

\begin{figure}[ht]
  \centering
  \begin{subfigure}{0.4\textwidth}
    \includegraphics[width=\linewidth]{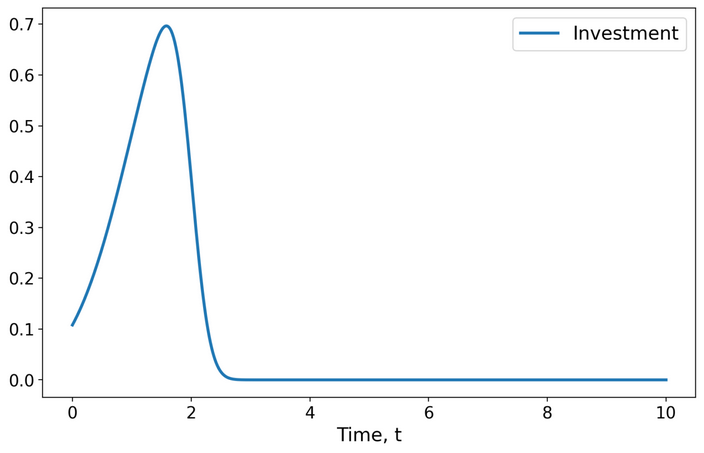}
    \caption{Investment in new tests}
    \label{fig:market_dynamics_panel_sub1}
  \end{subfigure}
  \begin{subfigure}{0.4\textwidth}
    \includegraphics[width=\linewidth]{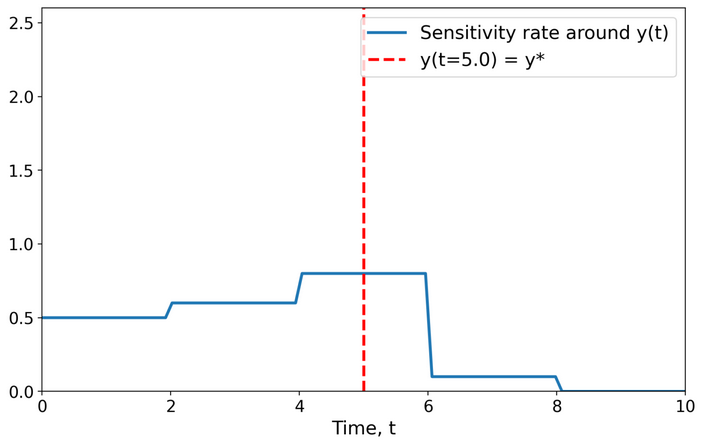}
    \caption{Test sensitivity rates around $y(t)$ at time $t$}
    \label{fig:market_dynamics_panel_sub2}
  \end{subfigure}
  \begin{subfigure}{0.4\textwidth}
    \includegraphics[width=\linewidth]{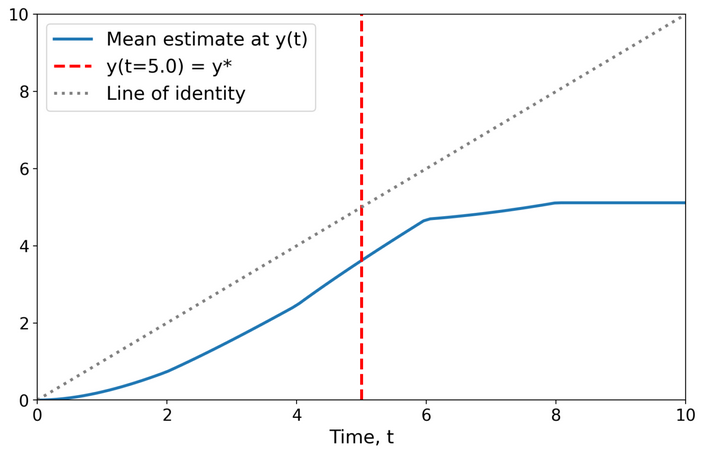}
    \caption{Danger bias at time $t$}
    \label{fig:market_dynamics_panel_sub3}
  \end{subfigure}
  \begin{subfigure}{0.4\textwidth}
    \includegraphics[width=\linewidth]{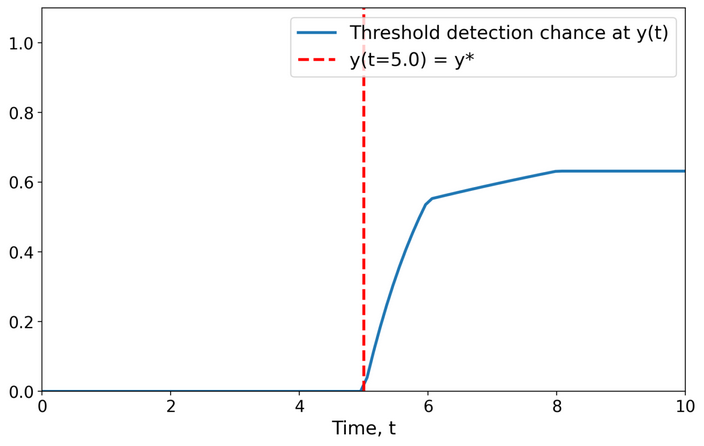}
    \caption{Threshold detection likelihood at time $t$}
    \label{fig:market_dynamics_panel_sub4}
  \end{subfigure}
  \caption{Market dynamics lead to growing bias in tracking AI dangers over time and longer threshold detection lags: Initial investment in new tests of novel risks falls quickly in response to market pressure. Eventually AI systems overtake the early-funded tests in exhibiting dangerous capabilities. Initial funding allows tests to be accurate for a while, after which bias and detection lags grow quickly.}
  \label{fig:market_dynamics_panel}
\end{figure}

Competition between AI labs encourages AI labs to keep the allotted time to model evals low. AI models benefit enormously from network effects: the more users the system has, the more likely others will choose your product over a competitor. Releasing your product first can therefore offer a large first-mover advantage. The ability to infer a plausible release schedule for your rivals may also induce a soft deadline. Clearly, these incentives strongly pressure AI labs to keep the time allotted to model evaluations short and predictable.

Initially, AI labs may be willing to support or conduct extensive evaluations. However, in the absence of third-party evaluations, we anticipate that the investment in new tests decreases with increasing pressure. Figure~\ref{fig:market_dynamics_panel_sub1} illustrates how initial investments fall rapidly as the capabilities of AI systems (and the rewards available to developers) increase.

This leads to a reduction in test sensitivity at higher levels of severity. Note that the axes in Figure~\ref{fig:market_dynamics_panel_sub2} are different from those in the previous figures. On the y-axis, we plot the test sensitivity rates near the hidden level of danger at time $t$, $y(t)$, while the x-axis tracks the time, $t$. As the test suite is being built over time, plotting our results in this way lets us focus only on the tests that provide new information over time. The first observation we make is that tests for greater severity dangers are worst affected by rapidly declining investments. This has important consequences for the bias in estimating danger. Initially, the bias was shrinking. As the initial investments in new tests were large, this provided some buffer in constructing tests for upcoming levels of danger. But eventually the shortage in funding leads to a sudden plunge in the test sensitivities, so a large bias emerges just past the threshold; the lack of recovery in test investment results in an abysmal likelihood of detecting the threshold crossing at all, never mind with a lag. So, this s-shaped trend in bias may be especially dangerous if society is misled into thinking the dangers are well understood.

The above market dynamics are not limited to pre-deployment checks. Post-deployment, market incentives are still in play: competition encourages AI labs to rush to create more capable AI systems faster.
The resulting consequence is that the time available to build tests for high-severity systems may be substantially shorter than in the absence of such competition.\footnote{One advantage for testing is that the model is immediately accessible for testing by a range of actors, without the pressing time constraints induced by competitive pressure. However, this does not necessarily mean that enough resources go to the most informative tests. Even after release, AI labs are likely to continue competing. One dimension to compete on is how safe their systems appear. Yet, this can motivate tests which focus on risks that tend to fall, rather than rise, with capabilities \citep{ren2024safetywashing}. For those teams who are motivated to focus on evaluating systemic risks from AI systems, they will likely have to balance further evaluations with preparing to evaluate future systems, given that future deadlines for pre-deployment tests are likely to be strict. Overall, it is hard to make a case that post-deployment testing is likely to lead to quick revisions of estimates of danger (until surprising and concrete hazards attributable to the new system occur).} If higher severity evals require more time to design and implement,  then reducing the time available to build them harms dangerous capability evaluations.

There is also a subtler issue caused by market dynamics: If competition means that we sample relatively few jumps in AI capabilities, we will be less informed about the relationship between AI capabilities and danger. When our estimates are biased downward, as in Figure~\ref{fig:market_dynamics_panel_sub3}, infrequent sampling of AI capabilities can lead to sudden surprise spikes in bias, even beyond a danger threshold.

\subsubsection{Technical challenges to testing for greater danger}

There are several barriers to achieving high test sensitivity when evaluating higher levels of dangerous capabilities.
\begin{itemize}
    \item Model mispecification: our measures of danger are only proxies for the latent capabilities we are concerned about.
    \item Specification gaming: we should anticipate that if we tend to inhibit behaviours from models which we see as dangerous, that we are more likely to end up with models who have similar dangerous capabilities but are less detectable \citep{krakovna2020specification}.
    \item Model deception: AI systems may selectively underperform when they are aware that they are being assessed for potential dangers as being seen as dangerous makes completing their implicit goals more difficult. This may or may not require deceptive capabilities \citep{benton2024sabotage}.
\end{itemize}

At first glance, it would appear that the effect on test sensitivity over time is similar to Figure~\ref{fig:market_dynamics_panel}. However, this understates the issue. The above challenges mean that it is quite likely that the usefulness of existing indicators will degrade over time. This means that it is a challenge not only to detect higher levels of danger but also to stay on top of the ways in which current dangers manifest. Were the erosion of test effectiveness to be combined with the challenges we previously discussed, it would be prudent to anticipate that poorly designed tests could greatly mislead us into believing our systems to be getting safer when the opposite is true.

The above discussion assumes that dangerous capabilities become harder to detect for more capable AI systems. However, it is worth noting that the dynamics here can be non-linear. Before an AI system is widely competent at a skill such as deception, the AI system may only exhibit the dangerous capability in environments that are well suited for that skill. In such cases, a wider net may need to be cast to detect these intermediate levels of severity, suggesting that high test sensitivities may be difficult to achieve. As the AI system becomes adept at employing the skill in many settings, it may be much easier to find evidence of this capability. Test sensitivity rates may increase. However, we hinted above that AI systems may themselves learn to apply skills such as deception while they are evaluated. We may anticipate test sensitivities to fall again unless there is a targeted effort to find more robust ways to evaluate deceptive systems.

\subsection{How to build and not to build an AI testing framework over time}

Policy makers may wish to strategically target support for dangerous capability tests in the hopes of overcoming some of the barriers to testing effectiveness. 

We next cover a selection of useful insights from our simulation of building a testing suite to track dangerous capabilities. We find reasons to suggest that prioritising a single goal when investing in tests can backfire. We focus on two goals motivated by our model: (i) minimising the bias in tracking the severity of danger presented by AI systems, and (ii) maximising the likelihood of detecting a crossing of a specified danger threshold. We assume a scenario where resources for tests of a danger threshold compete with resources for tracking AI dangerous capabilities. This could be the case if the danger threshold is somewhat far from current estimates of danger and resources are limited so that achieving high test sensitivity for a wide range of danger levels is not possible.

\textbf{Finding 1:} If we focus limited resources on only close-to-frontier dangers, then our bias will consistently grow over time.

Consider a scenario of complete knowledge about the sensitivity rate of our tests and the resources needed to achieve this for future tests. We do not know the current danger level, but we have estimates of the past danger levels, which we will assume are biased downward to some degree. It does not help the first goal to pour resources into tests for higher levels of danger than necessary. However, it may be difficult to know for sure how quickly the danger level is rising with capabilities.

If we could accurately forecast how the level of danger rises with capabilities and how capabilities change over time, then the optimal strategy would be to place all resources on and just before that new danger level.

However, let us assume we are much more uncertain. We instead act as a Bayesian and form a posterior prediction about plausible levels of danger for upcoming AI systems. Then, optimally spreading resources is all about increasing the likelihood of capturing the more likely higher end of that distribution. As our inference on future danger depends on previous estimates of danger, we can run into several issues:
\begin{itemize}
    \item Biases will stack up over time, especially if we have too narrow an inference on the new level of danger
    \item We may be vulnerable to inflection points in danger if we anchor too heavily on past trends in our danger estimates, even if we have been successful so far on minimizing bias. This even leads to predicting the inflection point much later.
\end{itemize}


\textbf{Finding 2:} If we focus only on detecting the crossing of a specified danger threshold, our ability to track trends in AI dangers suffers, which can leave us unprepared for new advances in AI systems.

A very different dynamic is plausible if we instead prioritise a danger threshold. If the threshold is reasonably far away, then we may not allow ourselves enough opportunity to see an upward trend towards the danger threshold. This is problematic for two reasons:
\begin{itemize}
    \item We may set a danger threshold too high - if harms are possible before the danger threshold, then maximising for goal 2 will allow harms to occur that could have been spotted in advance under a more flexible strategy. This may especially be true if the combination of different dangerous capabilities can be more severe than the same level of each danger on its own. If cascade risks are plausible, that is, attaining the capability quickly implies unlocking other dangerous capabilities or paths to harm, then this is even more of an issue.
    \item We may need time to set up effective research and policy streams in response to crossing the danger threshold. This is also true when supporting more effective capability evaluations. If we do not observe AI systems approaching this threshold, we may be more likely to underinvest in the tests themselves, or even reduce investment and encourage even faster progress under the false security of seeing little evidence that the danger threshold is near.
\end{itemize}

This leads us to our first takeaway. 

\textbf{Takeaway 1:} A fixed per time-step budget should balance investment in higher severity tests with tests closer to the current estimated frontier. This allows consistent progress in tracking AI capabilities, while setting a ceiling on lags in threshold monitoring.

The non-triviality of building a high-quality test suite is concerning. We would not want this challenge to paralyse decision makers from making any decision.  Now for our second takeaway. 

\textbf{Takeaway 2:} The longer the delay in building a high-quality test suite, the greater the investment needed in a shorter time to even have a moderate chance of detecting danger.

To see the above, consider two issues: first, to achieve a similar detection rate as when we begin earlier, we need to invest more in the test suite in a shorter amount of time (assuming progress in AI systems is unchanged). Second, due to a delay in testing, we have much less information about the rate of increase in AI danger. It is therefore less likely that we correctly estimate where to start investing, which reduces our odds of detecting the danger early. Thus, delays in spending on high-quality test suites leave us in a poor position for estimating risks.




\subsection{Scenario planning}

We now demonstrate the applicability of our model as a tool for being more proactive in setting policy for frontier AI systems.

To achieve this, we have sketched a selection of scenarios that may be relevant to future decision makers in AI. We then say what one can probably infer about the bias or threshold monitoring lag and, where relevant, identify key bottlenecks to better testing. We then briefly sketch out a potential policy response.

For these scenarios, we will assume a risk-averse government who would be unwilling to accept even moderate risks of catastrophe and so wishes to avoid significant delays in detecting dangerous capabilities above their chosen threshold.

\begin{itemize}
    \item  Scenario 1
    \begin{itemize}
        \item Context: New capabilities appear safe. Evidence that safety testing was limited. Minor concerns now reported.
    \end{itemize}
    \begin{itemize}
        \item  Inference: Risks underestimated, perhaps with a large bias and lag.
    \end{itemize}
    \begin{itemize}
        \item Policy Response: Invest significantly in existing and novel testing methodologies to counter existing underinvestment. Target ahead of current reported issues. Mandate current best practise in AI Safety testing.
    \end{itemize}
    \item Scenario 2
    \begin{itemize}
        \item Context: So far testing has revealed that AI systems continue to present more dangerous capabilities. Pace is slow but consistent. Capabilities are not yet near any threshold, and the behaviours do not cause any incidents in practise.
    \end{itemize}
    \begin{itemize}
        \item Inference: Dangerous capabilities may be well estimated.
    \end{itemize}
    \begin{itemize}
        \item Policy Response: Invest in novel testing methodologies at or near the threshold. Pre-register these tests with hypotheses about what results we'd see if current estimates are accurate and results we'd see if current estimates are biased. Reallocate resources given these results.
    \end{itemize}
    \item Scenario 3
    \begin{itemize}
        \item Context: As in scenario 2, but new tests indicate much higher growth in several dangerous capabilities.
        \item Inference: Possible inflection point in relationship between AI progress and dangerous capabilities (or previous tests underestimated dangerous capabilities)
        \item Policy Response: Increase mandates for extensive safety testing, including follow-up testing. The value of new information is extremely high due to the candidate inflection point.
    \end{itemize}
\end{itemize}

The next set of scenarios involve policy responses that go beyond simply improving the tests themselves. Instead, the focus is on direct policy levers that decision makers may wish to employ to reduce risks from AI systems when their testing framework leaves them with imperfect information. These scenarios are also more complex, so we devote more time to explaining the scenario and a potential response.

\begin{itemize}
    \item  Scenario 4
    \begin{itemize}
        \item Context: Past trends indicate that AI systems present risks that are initially underestimated upon release and are beginning to approach the chosen threshold. New AI systems which are more capable are expected to be released imminently.
    \end{itemize}
    \begin{itemize}
        \item  Inference: New AI systems may pose unacceptable risks but they could go undetected.
    \end{itemize}
    \begin{itemize}
        \item Policy Response: Consider buying more time for AI testing and safety research by mandating incremental scaling of AI systems. Consider the coordinated pausing framework in \citet{alaga2023coordinated} if international governance is needed.
    \end{itemize}
    \item Scenario 5
    \begin{itemize}
        \item Context:  New tests for AI capabilities are funded less. Progress in AI systems has slowed relative to their high rates of growth when large language models first achieved success. However, AI systems continue to grow more capable in small or somewhat larger steps. AI companies are anticipated to continue scaling their investments in training new AI systems. AI systems by the end of the decade are anticipated to be trained on several orders of magnitude more compute in only a few years. A substantial portion of these resources go towards AI approaches that have yet to see large benefits from scaling (such as methods based primarily on reinforcement learning where there is precedence for large and sudden capability jumps in narrow domains). 
    \end{itemize}
    \begin{itemize}
        \item Inference: Were AI tests to remain well-funded, you would expect to see a smooth trend in dangerous capabilities. It is hard to tell if diminishing increases in danger on the margin reflect the true relationship between capabilities and danger, or if this is driven by limited testing. Even if current funding captures the true trend, it is unclear if this will be sufficient to track the dangers if investments unlock new capability advances in AI.
    \end{itemize}
    \begin{itemize}
        \item Policy Response: Prioritise safety testing for current and novel AI systems. Recognise that new capability advances are likely to only be investigated well if there is a strong existing ecosystem for monitoring risks from AI systems.
    \end{itemize}
    \item Scenario 6
    \begin{itemize}
        \item Context: AI systems are rapidly transforming the economy, due to the automation of large sectors in the economy or domains of scientific research. This is enabled by strong capabilities for long-horizon planning and acting autonomously in real-world settings. Data indicates that any current hazards from AI systems do substantially less damage than the enormous benefits from AI systems. Admittedly, the science of AI alignment is still in its infancy. Experts continue to raise concerns about dangerous capabilities suspected to be necessary for a future loss-of-control scenario, but there is limited research on how to assess for their presence.
        \item Inference: The assessed relationship between AI capabilities and AI harms has so far surprised experts in the field. So has the pace of growth in AI capabilities. However, it is apparent that the quality of testing has rapidly declined (otherwise, there would be much less uncertainty as to what extent these very capable AI systems exhibit the most severe dangerous capabilities). It is reasonable to expect AI systems to have already passed specified thresholds, especially if AI systems prove capable of strategically underperfoming in simple evaluations.
        \item Policy Response: Direct resources to testing for the most severe dangerous capabilities. Restrict critical systems in infrastructure and decision making from adopting AI systems (or add human-operated mechanisms as a check against critical systems that have already been automated by AI). Demand strong safety cases (or preferably guarantees) before automating any critical system, and do so incrementally. Support an international initiative to coordinate the testing of AI systems and reach agreement on red lines for AI use (to reduce incentives for risk taking with new applications of AI). 
    \end{itemize}
\end{itemize}


\subsection{Towards a case study}
\label{sec:case_study}

It is outside the scope of the current work to release a full case study that makes use of our model. In short, it is challenging to infer the test sensitivity rates implied by real-world evaluations. We aim to make these challenges clear by selecting a few examples of recent work in the field.

Apollo Research released a report on their evaluations for in-context scheming across many frontier language models \citep{benton2024sabotage}.\footnote{The scheming is considered "in-context" because the goal they were instructed to prioritise (which we later see them scheme to maintain) is given to them in their context window (e.g. by including the goal in a system message when prompting the model).} Previously, these results were only available for Open AI's o1-preview but now extend to o1's full release, as well as Anthropic, Meta, and Google Deepmind's own frontier models. Scheming is defined as the pursuit of misaligned objectives through deceptive means. Two examples they observe are attempts to exfiltrate their model weights (often lying about these attempts even after multiple turns of questioning) and strategically underperforming on dangerous capability evaluations. In short, in-context scheming is widely observed across all language models, and there is evidence that models engage in scheming behaviours even without any strong nudge to stick to their objective, suggesting that some of these behaviours are learnt during training. These evaluations typically report the percentage of responses that exhibited scheming behaviour, aggregated across variant tasks, but separated across six task families. 

METR has also published findings regarding their software automation evaluations \citep{kinniment2024evaluating}.  Their resources for pre-deployment testing were much lower than what their example protocol indicated as conservative \citep{r2024example, i2024openai}. They had 2 evaluation test suites and tested an o1 model assisted by a set of different agent scaffolds. Their first test suite covered a set of 50 task families related to autonomous research capabilities. They reported the pass rates for each task family as well as the average pass rate and their standard errors. These standard errors should be understood as reflecting variability in the sampling of different tasks: had they chosen a different combination of similar tasks, they would anticipate the results to be within these standard errors. The second evaluation suite focused on a set of difficult ML research workflows. As these tasks are challenging to complete, the use a milestone indicator to track progress towards completing each task: the more subtasks they complete, the higher a score they receive for that task even though they fail to complete it. \footnote{They note that they expect these results to be underestimates given the limited compute time for the agents, and the limited time to iterate on more effective agent scaffolds to allow o1 to tackle the tasks \citep{r2024details}.}

It is true that there are similarities in the case studies above to our model. Different evaluations are clearly meant to prioritise risks of different severity. Completion of METR's ML research tasks would indicate a riskier model than one which could only perform well on their small set of generic software research tasks. \footnote{These similarities are not too suprising since METR acknolwedges that they aim to ensure more continuous tracking of AI capabilities rather than focussing only on detecting specific milestones \citep{r2024example}.} It is also clear from reading the report from Apollo Research that while they are confident in their ability to detect in-context scheming, there is much work to be done to be confident in evaluating for more general scheming. 

However, noticing these similarities is not sufficient to map their results to the test sensitivity rate defined in our model. The model we presented made some relatively strong assumptions that allow us to illustrate how it could look to be failing or succeeding in having a testing framework for dangerous capabilities. These assumptions are unlikely to hold in practise.

The above case studies make clear that:
\begin{enumerate}[label=(\roman*)]
    \item  we usually will want a composite indicator which reflects the use of a sequence of evaluation suites for tracking AI capabilities. This is unlikely to be well captured by our piecewise-constant test sentitivity rates;
    \item  we usually work with test suites and estimators where false positives are plausible. In addition, the rate of false negatives for any given danger score may grow or shrink with the current capabilities of frontier AI systems. Relaxing this assumption means we need to be more careful with tracking how our estimator for the level of danger changes over time (it's no longer as simple as censoring the right of the distribution);  and
    \item  we will usually want to supplement the data reported by evaluators with additional data from experiments testing the calibration of each individual estimator used in their work.
\end{enumerate}

Given the above challenges, we leave the development of a more complete methodology to future work.

\section{Open questions}
\label{sec:questions}

We propose the following open questions and welcome researchers to work on them.

\begin{itemize}
    \item How can we build a complete case study applying our model, as discussed in \ref{sec:case_study}?
    \item What does an equivalent model of an upper-bound estimator tell us about our ability to track AI dangers? How should auditors weigh the value of both estimators? This is useful because such an estimator may be useful in helping us rule out levels of danger that a model cannot yet achieve. We anticipate that this idea requires only a minor variation of the approach we take in this work.
    \item Our model may be helpful in identifying perverse incentives to engage in low-quality auditing. If firms or governments allocate funding to tests that are unlikely under our model to reduce the bias or detection lags for dangerous capabilities, this may be an indicator that the incentives are not having their desired effect. Further work should connect our model to examples of incentives used in practise.
    \item Can we extend the model to capture the relationships between multiple continuous estimators of AI danger? This is important for better assessing the effectiveness of pre-deployment tests for AI systems because the greatest risks from AI systems come from agents that have a combination of dangerous capabilities (e.g., long horizon planning, scheming, and the ability to collude with others \citep{motwani2024secret}). If we know that some estimators are related, this also helps us better judge the value of information when making marginal changes to tests for each types of risk.
\end{itemize}

\newpage

\section*{Acknowledgements}

We would like to thank our anonymous reviewers for valuable feedback on our work.

\section*{Conflicts of Interest}

On behalf of all authors, the corresponding author states that there is no conflict of interest.

\section*{Data Availability}

Data and code for the simulations can be made available upon request.

\section*{CRediT authorship contribution statement}

\begin{itemize}
    \item Paolo Bova: Conceptualization of this study, Methodology, Software, Writing - Original Draft, Formal analysis,  Visualization
    \item Alessandro Di Stefano: Writing - Review \& Editing, Supervision
    \item The Anh Han: Writing - Review \& Editing, Supervision
\end{itemize}

\newpage

\setcounter{table}{0}
\renewcommand{\thetable}{A\arabic{table}}%
\setcounter{figure}{0}
\renewcommand{\thefigure}{A\arabic{figure}}%
\setcounter{section}{0}
\renewcommand{\thesection}{A\arabic{section}}%

\section*{Appendix}

\section{Derivation of CDF of estimator}
\label{sec:proof1}

\subsection{Problem Setup}
    
Consider a continuum of tests on the interval $[y_0, y_{\text{max}}]$. Let $Y$ denote the maximum of all tests that pass on this continuum. For any given test, $y$, we define the rate of passing the test as $r(y)$.

$r(y)$ can also be framed as the conditional rate of $Y=y$ being the maximum passing test, given the maximum is at most $y$ (assuming failing all greater tests gives no information about the likelihood of passing test $y$, these two ideas are equivalent).

\subsection{Derivation}

\subsubsection{Likelihood Function}

The likelihood that any particular test $y$ is the maximum of all passing tests on the continuum is given by:

\[
f(Y=y) = r(y) \cdot \Pr(Y \leq y) = r(y) \cdot F(y)
\]

where $f(Y=y)$ is the probability density function (PDF) and $F(y)$ is the cumulative distribution function (CDF).

\subsubsection{Differential Equation}

This relationship leads to a separable ordinary differential equation:

\[
\frac{f(y)}{F(y)} = r(y)
\]

\subsubsection{Solving the ODE}

We can rewrite this as:

\[
\frac{d}{dy} \ln F(y) = r(y)
\]

Integrating both sides:

\[
\int_{y}^{y_{\text{max}}} \frac{d}{du} \ln F(u) \, du = \int_{y}^{y_{\text{max}}} r(u) \, du
\]

This yields:

\[
\ln F(y_{\text{max}}) - \ln F(y) = \int_{y}^{y_{\text{max}}} r(u) \, du
\]

\subsubsection{Final CDF Expression}

Since $F(y_{\text{max}}) = 1$, we have:

\[
F(y) = \exp\left(-\int_{y}^{y_{\text{max}}} r(u) \, du\right)
\]

\subsubsection{PDF Derivation}

To derive the PDF, we differentiate the CDF. Let $R(y) = \int_{y_0}^y r(u) \, du$. Then we can rewrite the CDF as:

\[
F(y) = \exp(-(R(y_{\text{max}}) - R(y)))
\]

Differentiating:

\[
f(y) = \frac{d}{dy}F(y) = \exp(-(R(y_{\text{max}}) - R(y))) \cdot r(y) = F(y) \cdot r(y)
\]

This confirms our original formulation and ensures that $f(y)$ is indeed non-negative.

\subsection{Boundary Conditions and Point Mass}

\begin{enumerate}
    \item $F(y_{\text{max}}) = 1$ is trivially satisfied.
    \item $F(y_0) > 0$ implies a point mass at $y_0$.
    \item The point mass at $y_0$, if it exists, is characterized by the jump discontinuity in $F(y)$ at $y_0$. Specifically:
   \[
   P(Y = y_0) = F(y_0) = \exp\left(-\int_{y_0}^{y_{\text{max}}} r(u) \, du\right)
   \]
\end{enumerate}

\subsection{Reversed Hazard Rate}

From the expression for $F(y)$, we can see that $r(y)$ is indeed the reversed hazard rate (or accumulation rate) of $F(y)$:

\[
r(y) = \frac{d}{dy} \ln F(y) = \frac{f(y)}{F(y)}
\]

which is the definition of the reversed hazard rate.

\subsection{Conditions on the reversed hazard rate}

While continuity of $r(y)$ is not strictly necessary, we require:

\begin{enumerate}
    \item $r(y)$ must be non-negative for all $y \in [y_0, y_{\text{max}}]$.
    \item $r(y)$ must be integrable on $[y_0, y_{\text{max}}]$.
    \item $\int_{y}^{y_{\text{max}}} r(y) \, dy$ must be finite for all $y>y_0$.
\end{enumerate}

These conditions ensure that $F(y)$ is a valid CDF. Note that $r(y)$ can be defined in piecewise with jumps, as long as these conditions are satisfied.

\newpage

\printbibliography

@article{armstrong2016racing,
  title = {Racing to the Precipice: {{A}} Model of Artificial Intelligence Development},
  shorttitle = {Racing to the Precipice},
  author = {Armstrong, Stuart and Bostrom, Nick and Shulman, Carl},
  year = 2016,
  month = may,
  journal = {AI \& SOCIETY},
  volume = 31,
  number = 2,
  pages = {201--206},
  doi = {10.1007/s00146-015-0590-y},
  issn = {1435-5655},
  langid = {english}
}

@article{askell2019role,
  title = {The {{Role}} of {{Cooperation}} in {{Responsible AI Development}}},
  author = {Askell, Amanda and Brundage, Miles and Hadfield, Gillian},
  year = 2019,
  month = jul,
  journal = {arXiv},
  doi = {10.48550/arXiv.1907.04534},
  eprint = {1907.04534},
  eprinttype = {arxiv},
  primaryclass = {cs},
  archiveprefix = {arXiv}
}

@misc{alaga2023coordinated,
  title = {Coordinated pausing: An evaluation-based coordination scheme for frontier AI developers},
  author = {Jide Alaga and Jonas Schuett},
  year = 2023,
  url = {https://arxiv.org/abs/2310.00374},
  eprint = {2310.00374},
  archiveprefix = {arXiv},
  primaryclass = {cs.CY}
}

@misc{bengio2024international,
  title = {International {{Scientific Report}} on the {{Safety}} of {{Advanced AI}} ({{Interim Report}})},
  author = {Bengio, Yoshua and Mindermann, S{\"o}ren and Privitera, Daniel and Besiroglu, Tamay and Bommasani, Rishi and others},
  year = 2024,
  month = nov,
  publisher = {arXiv},
  number = {arXiv:2412.05282},
  doi = {10.48550/arXiv.2412.05282},
  url = {http://arxiv.org/abs/2412.05282},
  urldate = {2024-12-18},
  eprint = {2412.05282},
  primaryclass = {cs},
  archiveprefix = {arXiv}
}

@article{bengio2024managing,
  title = {Managing Extreme {{AI}} Risks amid Rapid Progress},
  author = {Bengio, Yoshua and Hinton, Geoffrey and Yao, Andrew and Song, Dawn and Abbeel, Pieter and others},
  year = 2024,
  month = may,
  journal = {Science},
  publisher = {American Association for the Advancement of Science},
  volume = 384,
  number = 6698,
  pages = {842--845},
  doi = {10.1126/science.adn0117},
  url = {https://www.science.org/doi/abs/10.1126/science.adn0117},
  urldate = {2024-12-18},
  copyright = {Copyright {\copyright} 2024 The Authors, some rights reserved; exclusive licensee American Association for the Advancement of Science. No claim to original U.S. Government Works},
  eprint = {https://www.science.org/doi/pdf/10.1126/science.adn0117},
  langid = {english}
}

@article{benton2024sabotage,
  title = {Sabotage {{Evaluations}} for {{Frontier Models}}},
  author = {Benton, Joe and Wagner, Misha and Christiansen, Eric and Anil, Cem and Perez, Ethan and others},
  year = 2024,
  journal = {ArXiv},
  doi = {10.48550/arXiv.2410.21514},
  url = {https://www.anthropic.com/research/sabotage-evaluations},
  urldate = {2024-12-13},
  langid = {english}
}

@article{bova2024both,
  title = {Both eyes open: Vigilant Incentives help auditors improve AI safety},
  author = {Bova, Paolo and Di Stefano, Alessandro and Han, The Anh},
  year = 2024,
  journal = {Journal of Physics: Complexity},
  publisher = {IOP Publishing},
  volume = 5,
  number = 2,
  pages = {025009}
}

@misc{buhl2024safety,
  title = {Safety Cases for Frontier {{AI}}},
  author = {Buhl, Marie Davidsen and Sett, Gaurav and Koessler, Leonie and Schuett, Jonas and Anderljung, Markus},
  year = 2024,
  month = oct,
  publisher = {arXiv},
  number = {arXiv:2410.21572},
  doi = {10.48550/arXiv.2410.21572},
  url = {http://arxiv.org/abs/2410.21572},
  urldate = {2024-12-18},
  eprint = {2410.21572},
  primaryclass = {cs},
  archiveprefix = {arXiv}
}

@misc{cottier2023who,
  title = {Who is leading in AI? An analysis of industry AI research},
  author = {Ben Cottier and Tamay Besiroglu and David Owen},
  year = 2023,
  eprint = {2312.00043},
  archiveprefix = {arXiv},
  primaryclass = {cs.CY}
}

@misc{dalrymple2024towards,
  title = {Towards {{Guaranteed Safe AI}}: {{A Framework}} for {{Ensuring Robust}} and {{Reliable AI Systems}}},
  shorttitle = {Towards {{Guaranteed Safe AI}}},
  author = {Dalrymple, David "davidad" and Skalse, Joar and Bengio, Yoshua and Russell, Stuart and Tegmark, Max and others},
  year = 2024,
  month = jul,
  publisher = {arXiv},
  number = {arXiv:2405.06624},
  doi = {10.48550/arXiv.2405.06624},
  url = {http://arxiv.org/abs/2405.06624},
  urldate = {2024-12-18},
  eprint = {2405.06624},
  primaryclass = {cs},
  archiveprefix = {arXiv}
}

@misc{goemans2024safety,
  title = {Safety Case Template for Frontier {{AI}}: {{A}} Cyber Inability Argument},
  shorttitle = {Safety Case Template for Frontier {{AI}}},
  author = {Goemans, Arthur and Buhl, Marie Davidsen and Schuett, Jonas and Korbak, Tomek and Wang, Jessica and others},
  year = 2024,
  month = nov,
  publisher = {arXiv},
  number = {arXiv:2411.08088},
  doi = {10.48550/arXiv.2411.08088},
  url = {http://arxiv.org/abs/2411.08088},
  urldate = {2024-12-18},
  eprint = {2411.08088},
  primaryclass = {cs},
  archiveprefix = {arXiv}
}

@misc{gruetzemacher2023international,
  title = {An {{International Consortium}} for {{Evaluations}} of {{Societal-Scale Risks}} from {{Advanced AI}}},
  author = {Gruetzemacher, Ross and Chan, Alan and Frazier, Kevin and Manning, Christy and Los, {\v S}t{\v e}p{\'a}n and others},
  year = 2023,
  month = nov,
  publisher = {arXiv},
  number = {arXiv:2310.14455},
  doi = {10.48550/arXiv.2310.14455},
  url = {http://arxiv.org/abs/2310.14455},
  urldate = {2024-12-13},
  eprint = {2310.14455},
  primaryclass = {cs},
  archiveprefix = {arXiv}
}

@misc{h_ojmark2024analyzing1,
  title = {Analyzing {{Probabilistic Methods}} for {{Evaluating Agent Capabilities}}},
  author = {H{\o}jmark, Axel and Pimpale, Govind and Panickssery, Arjun and Hobbhahn, Marius and Scheurer, J{\'e}r{\'e}my},
  year = 2024,
  month = oct,
  publisher = {arXiv},
  number = {arXiv:2409.16125},
  doi = {10.48550/arXiv.2409.16125},
  url = {http://arxiv.org/abs/2409.16125},
  urldate = {2024-12-18},
  eprint = {2409.16125},
  primaryclass = {cs},
  archiveprefix = {arXiv},
  langid = {english}
}

@misc{hendrycks2022unsolved,
  title = {Unsolved Problems in ML Safety},
  author = {Dan Hendrycks and Nicholas Carlini and John Schulman and Jacob Steinhardt},
  year = 2022,
  url = {https://arxiv.org/abs/2109.13916},
  eprint = {2109.13916},
  archiveprefix = {arXiv},
  primaryclass = {cs.LG}
}

@misc{hendrycks2023overview,
  title = {An {{Overview}} of {{Catastrophic AI Risks}}},
  author = {Hendrycks, Dan and Mazeika, Mantas and Woodside, Thomas},
  year = 2023,
  month = oct,
  publisher = {arXiv},
  number = {arXiv:2306.12001},
  doi = {10.48550/arXiv.2306.12001},
  url = {http://arxiv.org/abs/2306.12001},
  urldate = {2024-12-18},
  eprint = {2306.12001},
  primaryclass = {cs},
  archiveprefix = {arXiv}
}

@misc{ho2023international,
  title = {International {{Institutions}} for {{Advanced AI}}},
  author = {Ho, Lewis and Barnhart, Joslyn and Trager, Robert and Bengio, Yoshua and Brundage, Miles and others},
  year = 2023,
  month = jul,
  publisher = {arXiv},
  number = {arXiv:2307.04699},
  doi = {10.48550/arXiv.2307.04699},
  url = {http://arxiv.org/abs/2307.04699},
  urldate = {2024-12-18},
  eprint = {2307.04699},
  primaryclass = {cs},
  archiveprefix = {arXiv}
}

@misc{i2024openai,
  title = {{OpenAI o1 System Card}},
  author = {OpenAI},
  year = 2024,
  url = {https://cdn.openai.com/o1-system-card-20241205.pdf},
  urldate = {2024-12-18}
}

@misc{institute2024pre1,
  title = {Pre-{{Deployment Evaluation}} of {{Anthropic}}'s {{Upgraded Claude}} 3.5 {{Sonnet}}},
  author = {{UK AI Safety Institute}},
  year = 2024,
  url = {https://www.aisi.gov.uk/work/pre-deployment-evaluation-of-anthropics-upgraded-claude-3-5-sonnet},
  urldate = {2024-12-18},
  langid = {english}
}

@misc{institute2024pre2,
  title = {Pre-{{Deployment Evaluation}} of {{Open AI}}'s {{o1 model}}},
  author = {{UK AI Safety Institute}},
  year = 2024,
  url = {https://www.aisi.gov.uk/work/pre-deployment-evaluation-of-openais-o1-model},
  urldate = {2024-12-18},
  langid = {english}
}

@misc{kinniment2024evaluating,
  title = {Evaluating {{Language-Model Agents}} on {{Realistic Autonomous Tasks}}},
  author = {Kinniment, Megan and Sato, Lucas Jun Koba and Du, Haoxing and Goodrich, Brian and Hasin, Max and others},
  year = 2024,
  month = jan,
  publisher = {arXiv},
  number = {arXiv:2312.11671},
  doi = {10.48550/arXiv.2312.11671},
  url = {http://arxiv.org/abs/2312.11671},
  urldate = {2024-12-18},
  eprint = {2312.11671},
  primaryclass = {cs},
  archiveprefix = {arXiv}
}

@misc{koessler2024risk,
  title = {Risk Thresholds for Frontier {{AI}}},
  author = {Koessler, Leonie and Schuett, Jonas and Anderljung, Markus},
  year = 2024,
  month = jun,
  publisher = {arXiv},
  number = {arXiv:2406.14713},
  doi = {10.48550/arXiv.2406.14713},
  url = {http://arxiv.org/abs/2406.14713},
  urldate = {2024-12-18},
  eprint = {2406.14713},
  primaryclass = {cs},
  archiveprefix = {arXiv}
}

@misc{motwani2024secret,
  title = {Secret Collusion among Generative AI Agents},
  author = {Sumeet Ramesh Motwani and Mikhail Baranchuk and Martin Strohmeier and Vijay Bolina and Philip H. S. Torr and others},
  year = 2024,
  url = {https://arxiv.org/abs/2402.07510},
  eprint = {2402.07510},
  archiveprefix = {arXiv},
  primaryclass = {cs.AI}
}

@article{park2024ai,
  title = {{{AI}} Deception: {{A}} Survey of Examples, Risks, and Potential Solutions},
  shorttitle = {{{AI}} Deception},
  author = {Park, Peter S. and Goldstein, Simon and O'Gara, Aidan and Chen, Michael and Hendrycks, Dan},
  year = 2024,
  month = may,
  journal = {PATTER},
  publisher = {Elsevier},
  volume = 5,
  number = 5,
  doi = {10.1016/j.patter.2024.100988},
  issn = {2666-3899},
  url = {https://www.cell.com/patterns/abstract/S2666-3899(24)00103-X},
  urldate = {2024-12-18},
  langid = {english}
}

@misc{phuong2024evaluating,
  title = {Evaluating {{Frontier Models}} for {{Dangerous Capabilities}}},
  author = {Phuong, Mary and Aitchison, Matthew and Catt, Elliot and Cogan, Sarah and Kaskasoli, Alexandre and others},
  year = 2024,
  month = apr,
  publisher = {arXiv},
  number = {arXiv:2403.13793},
  doi = {10.48550/arXiv.2403.13793},
  url = {http://arxiv.org/abs/2403.13793},
  urldate = {2024-12-18},
  eprint = {2403.13793},
  primaryclass = {cs},
  archiveprefix = {arXiv}
}

@misc{r2023responsible,
  title = {Responsible {{Scaling Policies}} ({{RSPs}})},
  author = {METR},
  year = 2023,
  month = sep,
  journal = {METR Blog},
  url = {https://metr.org/blog/2023-09-26-rsp/},
  urldate = {2024-12-18},
  langid = {english}
}

@misc{r2024details,
  title = {Details about {{METR}}'s Preliminary Evaluation of {{OpenAI}} O1-Preview},
  author = {METR},
  year = 2024,
  month = sep,
  journal = {METR's Autonomy Evaluation Resources},
  url = {https://metr.github.io/autonomy-evals-guide/openai-o1-preview-report/},
  urldate = {2024-12-13},
  langid = {english}
}

@misc{r2024example,
  title = {Example Protocol},
  author = {METR},
  year = 2024,
  journal = {METR's Autonomy Evaluation Resources},
  url = {https://metr.github.io/autonomy-evals-guide/example-protocol/},
  urldate = {2024-12-13},
  langid = {english}
}

@misc{ren2024safetywashing,
  title = {Safetywashing: {{Do AI Safety Benchmarks Actually Measure Safety Progress}}?},
  shorttitle = {Safetywashing},
  author = {Ren, Richard and Basart, Steven and Khoja, Adam and Gatti, Alice and Phan, Long and others},
  year = 2024,
  month = jul,
  publisher = {arXiv},
  number = {arXiv:2407.21792},
  doi = {10.48550/arXiv.2407.21792},
  url = {http://arxiv.org/abs/2407.21792},
  urldate = {2024-12-18},
  eprint = {2407.21792},
  primaryclass = {cs},
  archiveprefix = {arXiv}
}

@article{reuel2024open1,
  title = {Open {{Problems}} in {{Technical AI Governance}}},
  author = {Reuel, Anka and Bucknall, Ben and Casper, Stephen and Fist, Tim and Soder, Lisa and others},
  year = 2024,
  month = jan,
  journal = {CoRR},
  url = {https://openreview.net/forum?id=GzdVEGq7Qj},
  urldate = {2024-12-18},
  langid = {english}
}

@misc{sevilla2024can,
  title = {Can AI Scaling Continue Through 2030?},
  author = {Jaime Sevilla and Tamay Besiroglu and Ben Cottier and Josh You and Edu Rold\'{a}n and others},
  year = 2024,
  url = {https://epoch.ai/blog/can-ai-scaling-continue-through-2030},
  note = {Accessed: 2024-12-18}
}

@article{t2024fact,
  title = {{{FACT SHEET}}: {{U}}.{{S}}. {{Department}} of {{Commerce}} \& {{U}}.{{S}}. {{Department}} of {{State Launch}} the {{International Network}} of {{AI Safety Institutes}} at {{Inaugural Convening}} in {{San Francisco}}},
  shorttitle = {{FACT SHEET}},
  author = {NIST},
  year = 2024,
  month = nov,
  journal = {NIST},
  url = {https://www.nist.gov/news-events/news/2024/11/fact-sheet-us-department-commerce-us-department-state-launch-international},
  urldate = {2024-12-18},
  langid = {english}
}

@article{cimpeanu2023social,
  title = {Social diversity reduces the complexity and cost of fostering fairness},
  author = {Theodor Cimpeanu and Alessandro {Di Stefano} and Cedric Perret and The Anh Han},
  year = 2023,
  journal = {Chaos, Solitons \& Fractals},
  volume = 167,
  pages = 113051,
  doi = {https://doi.org/10.1016/j.chaos.2022.113051},
  issn = {0960-0779},
  url = {https://www.sciencedirect.com/science/article/pii/S0960077922012309}
}

@article{han2020regulate,
  title = {To {{Regulate}} or {{Not}}: {{A Social Dynamics Analysis}} of an {{Idealised AI Race}}},
  shorttitle = {To {{Regulate}} or {{Not}}},
  author = {Han, The Anh and Pereira, Luis Moniz and Santos, Francisco C. and Lenaerts, Tom},
  year = 2020,
  month = nov,
  journal = {Journal of Artificial Intelligence Research},
  volume = 69,
  pages = {881--921},
  doi = {10.1613/jair.1.12225},
  issn = {1076-9757},
  copyright = {Copyright (c) 2020 Journal of Artificial Intelligence Research},
  langid = {english}
}

@article{han2022voluntary,
  title = {Voluntary Safety Commitments Provide an Escape from Over-Regulation in {{AI}} Development},
  author = {Han, The Anh and Lenaerts, Tom and Santos, Francisco C and Pereira, Lu{\'i}s Moniz},
  year = 2022,
  journal = {Technology in Society},
  publisher = {{Elsevier}},
  volume = 68,
  pages = 101843
}

@misc{jensen2023industrial,
  title = {Industrial Policy for Advanced AI: Compute Pricing and the Safety Tax},
  author = {Jensen, Mckay and Emery-Xu, Nicholas and Trager, Robert},
  year = 2023,
  publisher = {arXiv},
  doi = {10.48550/ARXIV.2302.11436},
  url = {https://arxiv.org/abs/2302.11436},
  copyright = {Creative Commons Attribution 4.0 International}
}

@misc{krakovna2020specification,
  title = {Specification Gaming: {{The}} Flip Side of {{AI}} Ingenuity},
  shorttitle = {Specification Gaming},
  author = {Krakovna, Victoria and Uesato, Jonathan and Mikulik, Vladimir and Rahtz, Matthew and Everitt, Tom and others},
  year = 2020,
  month = apr,
  note = {Retrieved February 2023 from https://deepmind.com/blog/article/Specification-gaming-the-flip-side-of-AI-ingenuity},
  langid = {english}
}

@misc{sevilla2023please,
  title = {Please {{Report Your Compute}}},
  author = {Sevilla, Jaime},
  year = 2023,
  month = apr,
  journal = {Epoch},
  urldate = {2023-06-01},
  howpublished = {https://epochai.org/blog/please-report-your-compute},
  langid = {english}
}

@misc{shevlane2023model,
  title = {Model Evaluation for Extreme Risks},
  author = {Shevlane, Toby and Farquhar, Sebastian and Garfinkel, Ben and Phuong, Mary and Whittlestone, Jess and others},
  year = 2023,
  month = may,
  publisher = {{arXiv}},
  number = {arXiv:2305.15324},
  doi = {10.48550/arXiv.2305.15324},
  urldate = {2023-05-29},
  eprint = {2305.15324},
  primaryclass = {cs},
  archiveprefix = {arxiv}
}

@article{whittlestone2021why,
  title = {Why and {{How Governments Should Monitor AI Development}}},
  author = {Whittlestone, Jess and Clark, Jack},
  year = 2021,
  month = aug,
  journal = {arXiv},
  doi = {10.48550/arXiv.2108.12427},
  eprint = {2108.12427},
  eprinttype = {arxiv},
  primaryclass = {cs},
  archiveprefix = {arXiv}
}

\end{document}